\documentclass[11pt]{article}

\usepackage[a4paper,margin=1in]{geometry}
\usepackage[utf8]{inputenc}
\usepackage[T1]{fontenc}
\usepackage[numbers,sort&compress]{natbib}
\usepackage{hyperref}
\usepackage{url}
\usepackage{booktabs}
\usepackage{amsfonts}
\usepackage{nicefrac}
\usepackage{microtype}
\usepackage{xcolor}
\usepackage{amsmath,amssymb}
\usepackage{graphicx}
\usepackage{subcaption}
\usepackage{multirow}
\usepackage{algorithm}
\usepackage{algpseudocode}
\usepackage{enumitem}

\title{LFHE: Local-First Heuristic Evolution for Bounded Local Topology Search in Decentralized Learning with Non-IID Data}

\author{
Yin-Kuan Liang$^{1}$ \quad Yan Gao$^{2}$ \quad Yang Long$^{1}$\\[0.5em]
$^{1}$ Durham University $^{2}$University of Cambridge
}
\date{October 2026}

\begin{document}

\maketitle

\begin{abstract}
Decentralized learning is highly sensitive to communication topology under non-IID data. Adaptive peer-selection methods can exploit local model information, but broader peer discovery may require increasingly large control state, whereas direct spectral optimization typically relies on graph-wide information. We study the intermediate setting of \emph{bounded local topology search} and propose \textbf{Local-First Heuristic Evolution (LFHE)}, a representation-driven rewiring framework whose candidate discovery and scoring use only ego-neighborhood and friend-of-a-friend (FoF) information. The structural score admits an exact interpretation through graph Dirichlet energy: its sum across clients equals twice the representation Dirichlet energy, which under standard linear consensus dynamics governs the instantaneous dissipation of representation disagreement. LFHE combines this state-dependent structural signal with early exploration and degree control, while algebraic connectivity remains an offline graph diagnostic. Under bounded sparse degree, its FoF candidate state remains local rather than expanding toward population-wide peer tracking. Across four image, speech, and text benchmarks, LFHE achieves competitive decentralized learning performance. Matched-protocol controls identify the structural term as the principal empirical topology-selection signal, while comparison with broader peer discovery exposes a trade-off between predictive performance and discovery-state locality. Together, these results motivate state-aware bounded local topology search between pairwise peer selection and globally informed topology optimization.
\end{abstract}

\section{Introduction}

Decentralized federated learning (DFL) enables collaborative model training without a central aggregation server. Instead, clients exchange and average model information over a peer-to-peer communication graph~\cite{koloskova2020decentralized,lian2017dpsgd}. Removing a central coordinator makes the communication graph part of the optimization dynamics: its connectivity and mixing properties strongly influence how quickly information propagates and local models move toward consensus \cite{boyd2006gossip,nedic2018topology,olfati2007consensus}.

The topology problem becomes more difficult under non-IID data. Sparse or poorly mixed graphs can sustain disagreement between local models and slow decentralized optimization~\cite{bellet2022dcliques,deVos2023epidemic,lebars2023refined}. Existing adaptive approaches often use model similarity or dissimilarity to choose communication partners~\cite{cox2026morph}. These signals are locally useful, but a sequence of locally attractive pairwise choices does not by itself determine whether the realized sparse graph supports efficient graph-wide information propagation. Conversely, methods that directly optimize mixing matrices or spectral graph objectives are more explicit about global communication structure, but commonly require graph-wide information or centralized computation~\cite{boyd2004fastest,xiao2004fast,olfati2007consensus}.

We study a middle ground: \emph{bounded local topology search}. Rather than reconstructing a global graph objective, each client searches only its ego-neighborhood and current FoF frontier. With maximum local degree \(\bar d\), the two-hop candidate state is \(O(\bar d^2)\) per client and \(O(N\bar d^2)\) system-wide, reducing to \(O(1)\) and \(O(N)\) under fixed sparse degree. By contrast, gossip-based discovery that progressively approaches population-wide peer coverage, as in Morph~\cite{cox2026morph}, can grow toward \(O(N)\) state per client and \(O(N^2)\) system-wide. This comparison concerns discovery/control state rather than total communication volume.

Based on this view, we propose \textbf{Local-First Heuristic Evolution (LFHE)}. At each topology update, a client samples an FoF candidate and evaluates an edge addition or one-edge swap using structural disagreement, early exploration, and degree control. The structural score is a node-local contribution to the representation Dirichlet energy; under standard linear consensus dynamics, this global energy governs the instantaneous dissipation of representation disagreement. LFHE does not compute algebraic connectivity \(\lambda_2\); spectral quantities remain offline diagnostics. The reference implementation serializes topology operations, so our decentralization claim concerns local information requirements rather than concurrent or asynchronous execution guarantees.

Our contribution is threefold. First, we formulate adaptive communication as bounded ego-neighborhood topology search, separating local online information from global offline diagnostics. Second, we establish an exact graph-theoretic interpretation of LFHE’s structural score: its node-wise values decompose the representation Dirichlet energy, which under standard linear consensus dynamics governs instantaneous disagreement dissipation. Third, we evaluate LFHE across image, speech, and text tasks using matched mechanism controls, sensitivity and communication analyses, update-order diagnostics, and comparison with broader peer-discovery strategies. The results identify the structural term as the principal topology-selection signal in the tested regime while exposing the trade-off between localized discovery and broader peer exposure.

\section{Related Work}

\textbf{Decentralized learning and heterogeneous data.}
Federated learning usually aggregates client updates through a server~\cite{mcmahan2017communication}, with methods such as FedProx~\cite{li2020fedprox} and SCAFFOLD~\cite{karimireddy2020scaffold} addressing optimization under heterogeneity. DFL instead relies on peer-to-peer communication~\cite{koloskova2020decentralized,lian2017dpsgd}. Classical distributed optimization and gossip results make clear that convergence and consensus depend strongly on the communication graph~\cite{nedic2009subgradient,tsitsiklis1986distributed,boyd2006gossip,nedic2018topology}. D-Cliques~\cite{bellet2022dcliques} and STL-FW~\cite{lebars2023refined} explicitly incorporate heterogeneity into sparse communication design, but rely on broader graph or mixing-matrix construction than the local online search considered here.

\textbf{Adaptive peer discovery and topology evolution.}
Epidemic Learning~\cite{deVos2023epidemic} uses randomized sparse communication. Morph~\cite{cox2026morph} performs gossip-based peer discovery and model-dissimilarity-driven replacement, progressively broadening discovered-peer state to enlarge peer-selection exposure. LFHE instead restricts candidate exposure to the current FoF frontier and evaluates the resulting ego-topology after an add or swap. The distinction is primarily in search scope: under bounded sparse degree, LFHE operates over a bounded local candidate frontier, whereas gossip-based peer discovery can progressively expose a client to a much broader portion of the network.\looseness-1

\textbf{Connectivity-oriented graph design.}
Graph-theoretic results show that locally constructed sparse graphs need not have favorable global connectivity, particularly as the system grows~\cite{balister2005knn,balister2009critical,xue2004neighbors}. More generally, decentralized averaging and consensus depend strongly on global graph and mixing properties~\cite{boyd2004fastest,olfati2004consensus,olfati2007consensus,sun2006fastest}. Direct optimization of graph structure or mixing matrices explicitly targets these properties~\cite{boyd2004fastest,xiao2004fast}, including in DFL-specific sparse overlays~\cite{hua2022overlay}, but typically requires graph-wide state or centralized optimization.

\textbf{Positioning.} LFHE occupies the space between pairwise peer-selection methods and direct global graph optimization. Its novelty is not dynamic topology adaptation itself, nor the use of model dissimilarity. It is the combination of restricted FoF candidate discovery, post-operation ego-topology scoring, and local add-or-swap updates without online global graph statistics. The structural score is state-dependent rather than a topology-only spectral objective: it is a node-local contribution to the representation Dirichlet energy, whose aggregate governs instantaneous disagreement dissipation under standard linear consensus dynamics, whereas \(\lambda_2\) characterizes the graph's worst-case normalized dissipation over non-consensus directions. LFHE does not estimate or optimize \(\lambda_2\).

\section{Method}

We formulate topology adaptation as a constrained local combinatorial search problem. At each topology-update step, client \(i\) restricts its candidate edge operations to those available from its ego-neighborhood and FoF frontier. LFHE performs sampled greedy search over this local action set and commits an operation only when it improves the initiator-local objective. The composite objective is a local decision rule, while its structural component admits an exact interpretation through representation Dirichlet energy and consensus dissipation.

\subsection{Problem Setup}

We consider a decentralized learning system with $N$ clients connected by a time-varying undirected graph
\[
G_t=(V,E_t),
\]
where node $i\in V$ denotes a client and edge $(i,j)\in E_t$ indicates that clients $i$ and $j$ can exchange model information at round $t$. Let $\mathcal{N}_i(t)$ denote the neighbor set of client $i$.

Each client $i$ holds a private local dataset $\mathcal{D}_i$ and maintains a local model $w_i^t$. At round $t$, client $i$ performs local training on $\mathcal{D}_i$ and then communicates only with neighbors in $\mathcal{N}_i(t)$. Its local objective is
\[
F_i(w)=\mathbb{E}_{(x,y)\sim \mathcal{D}_i}[\ell(w;x,y)],
\]
and the global learning objective is
\[
\min_{w} F(w)=\sum_{i=1}^{N} p_i F_i(w),
\]
where $p_i\ge 0$ and $\sum_{i=1}^{N} p_i=1$.

The communication graph strongly influences how efficiently information propagates across clients and how quickly local models approach consensus, especially under non-IID data. We therefore treat topology as an adaptive decision variable rather than a fixed background structure.

Our setting is subject to three operational constraints: during online topology adaptation, no client requires a global graph statistic; candidate discovery is restricted to ego-neighborhood and FoF information; and online edge acquisition is regulated by a degree threshold \(D_{\max}\). In the reference implementation, the initial connected Erd\H{o}s--R\'enyi graph is generated with expected degree four and may contain nodes whose initial degree exceeds \(D_{\max}\). Consequently, \(D_{\max}\) should be interpreted as an \emph{online degree threshold}: a new edge is added only when both endpoints are below the threshold, while a feasible swap preserves the initiator's degree.

Under these conditions, the objective is to perform bounded local topology search without global spectral computation. LFHE evaluates initiator-local improvement only; it does not require an affected peer endpoint to improve its own score, and it does not guarantee preservation or monotonic improvement of global connectivity.

\paragraph{Local topology-search problem.}
For client \(i\), let \(\mathcal{A}_i(G_t,k)\) denote the feasible one-step actions induced by sampled FoF candidate \(k\): keep \(G_t\), add \((i,k)\) when the degree condition holds, or replace an incident edge \((i,j)\) with \((i,k)\) when local redundancy and degree conditions permit. For feasible \(G'\), define
\begin{equation}
\label{eq:local_objective}
J_i(G';t)=
\omega_1(1-\beta_t)C_i(G')
+\omega_2\beta_t D_i(G')
-\omega_3\frac{\deg_{G'}(i)}{D_{\max}}.
\end{equation}
Given \(k\), LFHE selects the highest-scoring feasible local operation and commits it only if \(J_i(G';t)>J_i(G_t;t)\). Thus LFHE performs sampled greedy search over a locally observable topology neighborhood rather than optimization over the global graph space.

\subsection{Overview of the Framework}

LFHE performs online topology adaptation through local rewiring. At each topology-update round, client \(i\) samples a FoF candidate and tests a direct addition or one-edge swap. A proposal is committed only when it improves the initiator's score and satisfies the degree and local redundancy checks. Candidate discovery therefore remains local, while online edge acquisition is explicitly regulated.

The objective combines three signals with distinct roles. The structural term scores representation disagreement over incident communication edges; Section~\ref{sec:consensus_theory} shows that it is exactly a node-local contribution to representation Dirichlet energy. The mismatch term supplies early exploration, encouraging clients to examine alternative local neighborhoods rather than selecting dissimilar peers indiscriminately. The degree term regulates edge acquisition and discourages excessive communication concentration. Beyond the structural component, their combination remains a heuristic design choice whose role is evaluated empirically.

\begin{figure}[t]
    \centering
    \includegraphics[width=\linewidth]{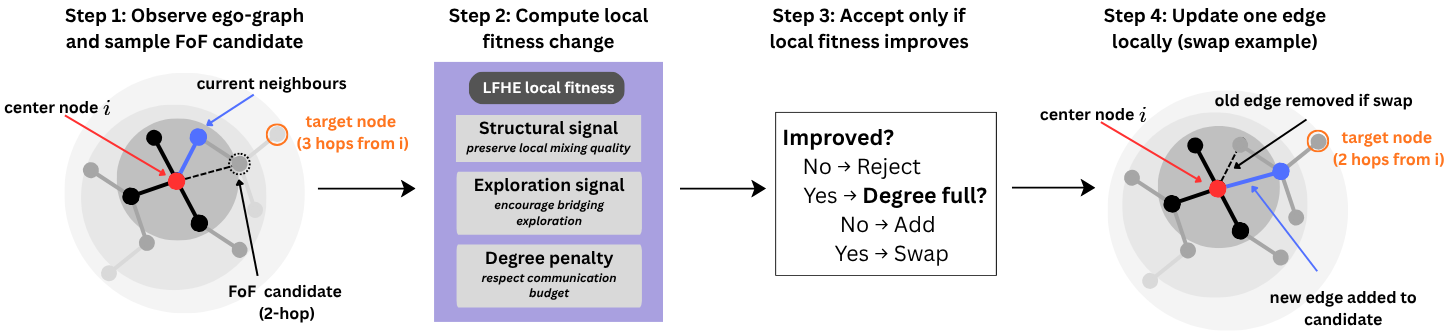}
    \caption{\small Overview of LFHE. Each client $i$ observes only its ego-graph, samples a two-hop FoF candidate, and evaluates it using a local objective with structural, exploration, and degree-control terms. If the candidate improves the objective, LFHE adds the edge when direct addition is feasible or performs a feasible one-edge swap otherwise; if no improving feasible operation exists, the candidate is rejected.}
    \label{fig:overview}
\end{figure}

As illustrated in Figure~\ref{fig:overview}, LFHE is a bounded local-search layer on top of decentralized learning. Each client observes only its local neighborhood, proposes a FoF add or swap, and accepts it according to initiator-local fitness. The reference experiments serialize proposals within a synchronous topology-update step; fully concurrent proposal conflicts, stale snapshots, and asynchronous commits are outside the evaluated protocol.

\subsection{Local Fitness Function Design}

The local objective in Eq.~\ref{eq:local_objective} uses \(\omega_1,\omega_2,\omega_3>0\), representation \(r_i\in\mathbb{R}^d\), and annealing coefficient
\[
\beta_t=\beta_0 e^{-\kappa t},
\]
where \(\beta_0>0\) controls the initial exploration weight and \(\kappa>0\) its decay. Representations are treated as fixed while candidate topology operations are evaluated. The three terms play distinct roles: the structural term targets current representation disagreement over incident communication edges, mismatch supplies early exploration, and degree regularization controls communication concentration.

\textbf{Local structural term.} \noindent
\begin{equation}
\label{eq:ci}
C_i(G) = \sum_{j \in \mathcal{N}_i(G)} \|r_i-r_j\|_2^2.
\end{equation}
If \(R=[r_1^\top,\ldots,r_N^\top]^\top\), then \(\frac12\sum_i C_i(G)=\mathrm{Tr}(R^\top L R)\), the representation Dirichlet energy of the current graph. Under standard linear consensus dynamics, this global energy equals the instantaneous rate at which centered representation disagreement is dissipated. Thus \(C_i(G)\) gives client \(i\) a locally computable contribution to a state-dependent consensus quantity without requiring a global graph statistic. It is not a surrogate for \(\lambda_2\): \(C_i\) depends on both the graph and the current representation field, whereas \(\lambda_2\) is a topology-only worst-case quantity.

\textbf{Neighborhood mismatch term.} \noindent
\begin{equation}
\label{eq:di}
D_i(G)
=
1 -
\cos\!\left(
r_i,\;
\frac{1}{|\mathcal{N}_i(G)|}
\sum_{j\in \mathcal{N}_i(G)} r_j
\right).
\end{equation}
This term measures directional mismatch between node \(i\) and its mean neighborhood representation. Unlike \(C_i(G)\), which has the Dirichlet-energy interpretation above, \(D_i(G)\) primarily supplies exploration pressure. Large values indicate directional mismatch with the current neighborhood and increase the influence of exploration early in training. It is therefore used to expose alternative local neighborhoods rather than to favor dissimilar peers indiscriminately.

\textbf{Degree regularization.} \noindent
The term \(\deg(i)/D_{\max}\) penalizes high initiator degree, discouraging excessive edge acquisition and communication concentration.

\textbf{Overall behavior.} \noindent
As \(\beta_t\) decays, the composite rule shifts from mismatch-driven exploration toward the consensus-grounded structural signal. Early in training, \(D_i(G)\) receives greater weight; later, \(C_i(G)\) becomes more influential. The annealing schedule and relative coefficients remain design choices; only the structural component carries the exact consensus-theoretic interpretation developed below.\looseness-1

\subsection{Topology Evolution Mechanism}

LFHE performs bounded local topology search using only ego-neighborhood and FoF information. For client \(i\), candidate peers are sampled from
\[
\Omega_i
=
\{ k \in \mathcal{N}_j \mid j \in \mathcal{N}_i,\ k \neq i,\ (i,k)\notin E \}.
\]
This two-hop construction expands peer exposure beyond the current neighborhood without requiring a global peer directory. Let \(d_i=|\mathcal{N}_i|\) and \(\bar d=\max_j d_j\). Before removing duplicates and already connected peers, the number of distinct two-hop candidates available to client \(i\) is bounded by
\[
|\Omega_i^{(2)}| \le d_i(\bar d-1).
\]
Hence the full FoF candidate space is \(O(\bar d^2)\) per client and \(O(N\bar d^2)\) system-wide. Under bounded sparse degree, this yields \(O(1)\) per-client and \(O(N)\) system-wide discovery state, so the local search space need not grow with population size. The implementation may sample only part of this frontier; these bounds characterize candidate discovery and control state rather than total communication volume.

For a sampled candidate, LFHE first considers direct addition when both endpoints satisfy the online degree threshold. Otherwise, it evaluates feasible one-edge swaps. A common-neighbor check preserves local redundancy when removing an incident edge, while accepted operations are selected according to the initiator-local objective.

Two structural properties follow. First, with representations fixed during a topology decision, every committed operation strictly improves the initiator objective \(J_i\). Second, each decision uses only ego/FoF structure, representations, and endpoint degree information, requiring neither a global peer directory nor online computation of \(L\) or \(\lambda_2\). Thus LFHE provides monotone local improvement within a population-independent per-client search space under bounded degree.

The reference implementation commits proposals sequentially within each topology-update step, so later clients observe earlier modifications and the realized graph can depend on processing order. The local redundancy check is not a global connectivity certificate, and improvement is guaranteed for the initiating client rather than every affected endpoint.

\subsection{Consensus-Dissipation Interpretation}
\label{sec:consensus_theory}

Let \(R=[r_1^\top,\ldots,r_N^\top]^\top\) stack client representations and let \(L=D-A\) denote the graph Laplacian. The structural scores exactly decompose the representation Dirichlet energy:
\begin{equation}
\mathcal{E}_G(R)
=
\mathrm{Tr}(R^\top L R)
=
\sum_{(i,j)\in E}\|r_i-r_j\|_2^2
=
\frac12\sum_i C_i(G).
\label{eq:dirichlet}
\end{equation}
Thus, \(C_i(G)\) is a node-local contribution to a global state-dependent consensus energy. Under standard linear consensus dynamics \(\dot R=-LR\), define centered representation disagreement as
\[
V(R)=\frac12\|PR\|_F^2,
\qquad
P=I-\frac1N\mathbf1\mathbf1^\top.
\]
Then
\begin{equation}
\dot V=-\mathcal{E}_G(R).
\label{eq:dissipation}
\end{equation}
Hence the representation Dirichlet energy is exactly the instantaneous rate at which centered disagreement is dissipated.

For fixed \(R\), replacing edge \((i,j)\) by \((i,k)\) changes this energy by
\[
\mathcal{E}_{G'}(R)-\mathcal{E}_{G}(R)
=
\|r_i-r_k\|_2^2-\|r_i-r_j\|_2^2.
\]
Therefore, a positive structural-energy change increases the instantaneous rate of disagreement dissipation for the current representation field.

Rayleigh--Ritz further gives
\begin{equation}
\mathcal{E}_G(R)
\ge
\lambda_2(L)\|PR\|_F^2.
\end{equation}
Thus, \(\lambda_2\) characterizes the graph's worst-case normalized dissipation over non-consensus directions, whereas LFHE's structural score evaluates the representation field currently realized on the graph. This establishes an exact consensus-theoretic interpretation for the structural component; the full LFHE objective additionally trades this signal against exploration and degree control and does not directly optimize \(\lambda_2\). The full derivation is provided in Appendix~\ref{appendix:theory}.

\subsection{Algorithm Description}

Each communication round consists of local training and decentralized aggregation, with an additional topology-adaptation step every $K$ rounds.

\textbf{Local training.}
Each client performs $E$ local epochs on its private dataset using the task-specific local optimizer.

\textbf{Decentralized aggregation.}
Each node aggregates neighbor models using degree-aware weights
\[
a_{ij} = \frac{1}{1 + \max(\deg(i), \deg(j))},
\qquad
a_{ii} = 1 - \sum_{j \in \mathcal{N}_i} a_{ij}.
\]

\textbf{Topology adaptation.}
Every $K$ rounds, each client samples a FoF candidate, evaluates the feasible addition and one-edge-swap operations induced by that candidate, and commits the highest-scoring feasible operation only when it improves the initiator-local objective. Full pseudocode and implementation details are provided in Appendix~\ref{app:algorithm}.

\section{Experimental Setup}
\label{sec:experimental_setup}

We evaluate LFHE on four datasets across three modalities: \textbf{CIFAR-10} and \textbf{CIFAR-100}~\cite{krizhevsky2009learning}, \textbf{Google Speech Commands}~\cite{warden2018speech}, and \textbf{Sentiment140}~\cite{go2009twitter}. Unless otherwise stated, data are partitioned with class-wise Dirichlet \(\alpha=0.1\), the main system has \(N=30\) clients, and LFHE uses online degree threshold \(D_{\max}=4\). The main comparison includes \textbf{FedAvg}~\cite{mcmahan2017communication}, \textbf{Fully Connected}, \textbf{Ring}~\cite{lian2017dpsgd}, \textbf{Static Random}, \textbf{Epidemic Learning}~\cite{deVos2023epidemic}, and \textbf{DissDL}, a legacy dissimilarity-guided implementation retained from our original benchmark.

For shared learning components, comparisons use matched data partitions, model initialization, local optimization budgets, and evaluation schedules where applicable. Topology-specific discovery, edge direction, and communication semantics follow each method's native design. Main results use five seeds \(\{42,43,44,45,46\}\); diagnostic studies use three seeds or a single controlled seed where explicitly stated. Complete model, optimizer, representation, initialization, and communication details are provided in Appendix~\ref{appendix:exp_details}. The connected Erd\H{o}s--R\'enyi initializer targets expected degree four and may contain nodes above \(D_{\max}\); accordingly, \(D_{\max}\) denotes an online edge-acquisition threshold rather than a hard maximum degree in the reference LFHE implementation.

\section{Main Convergence Results and Topology Analysis}

\begin{figure}[t]
\centering
\begin{minipage}{0.49\linewidth}
    \centering
    \includegraphics[width=\linewidth]{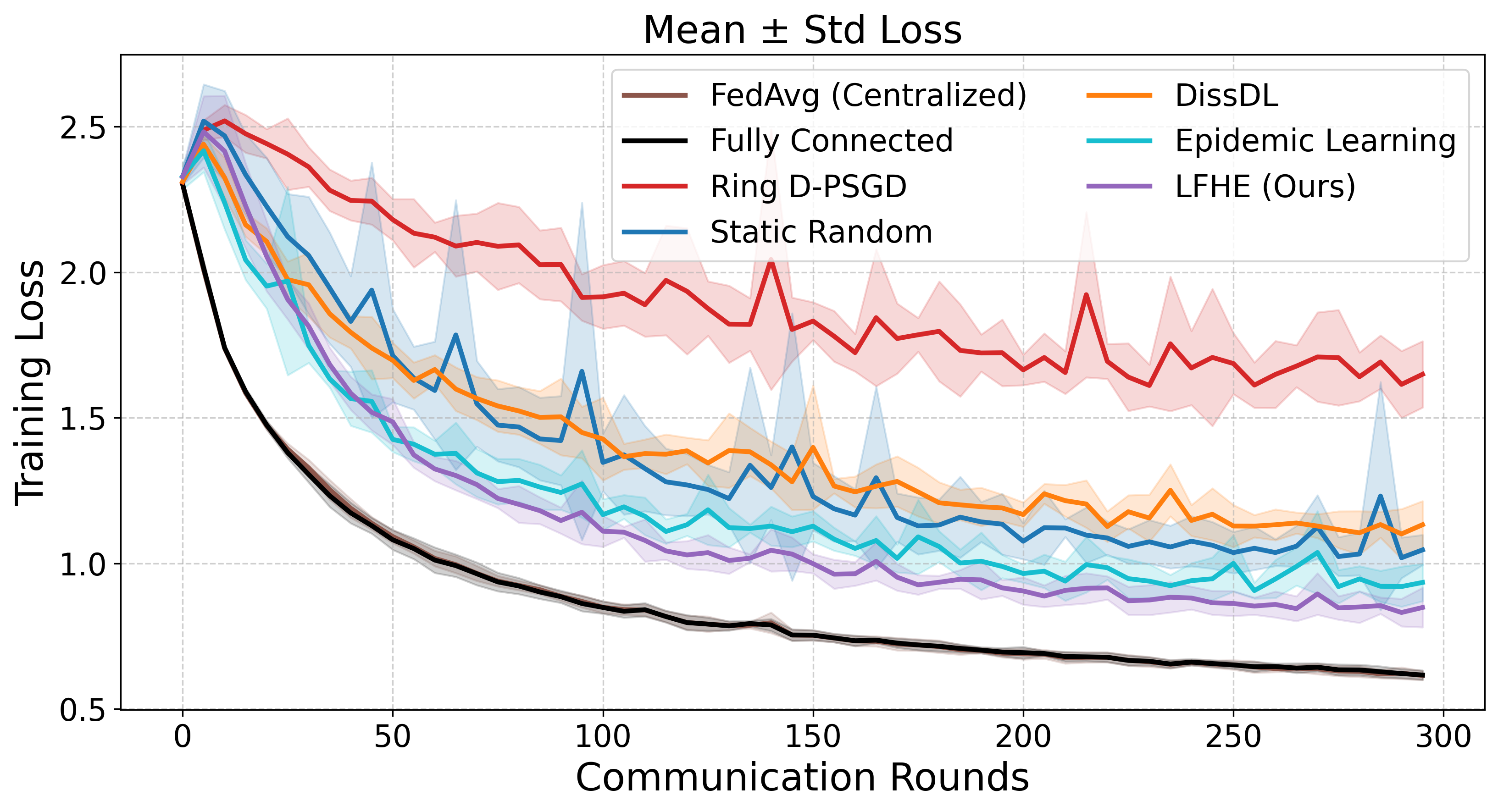}
    \caption*{\small(a) Training loss}
\end{minipage}
\hfill
\begin{minipage}{0.49\linewidth}
    \centering
    \includegraphics[width=\linewidth]{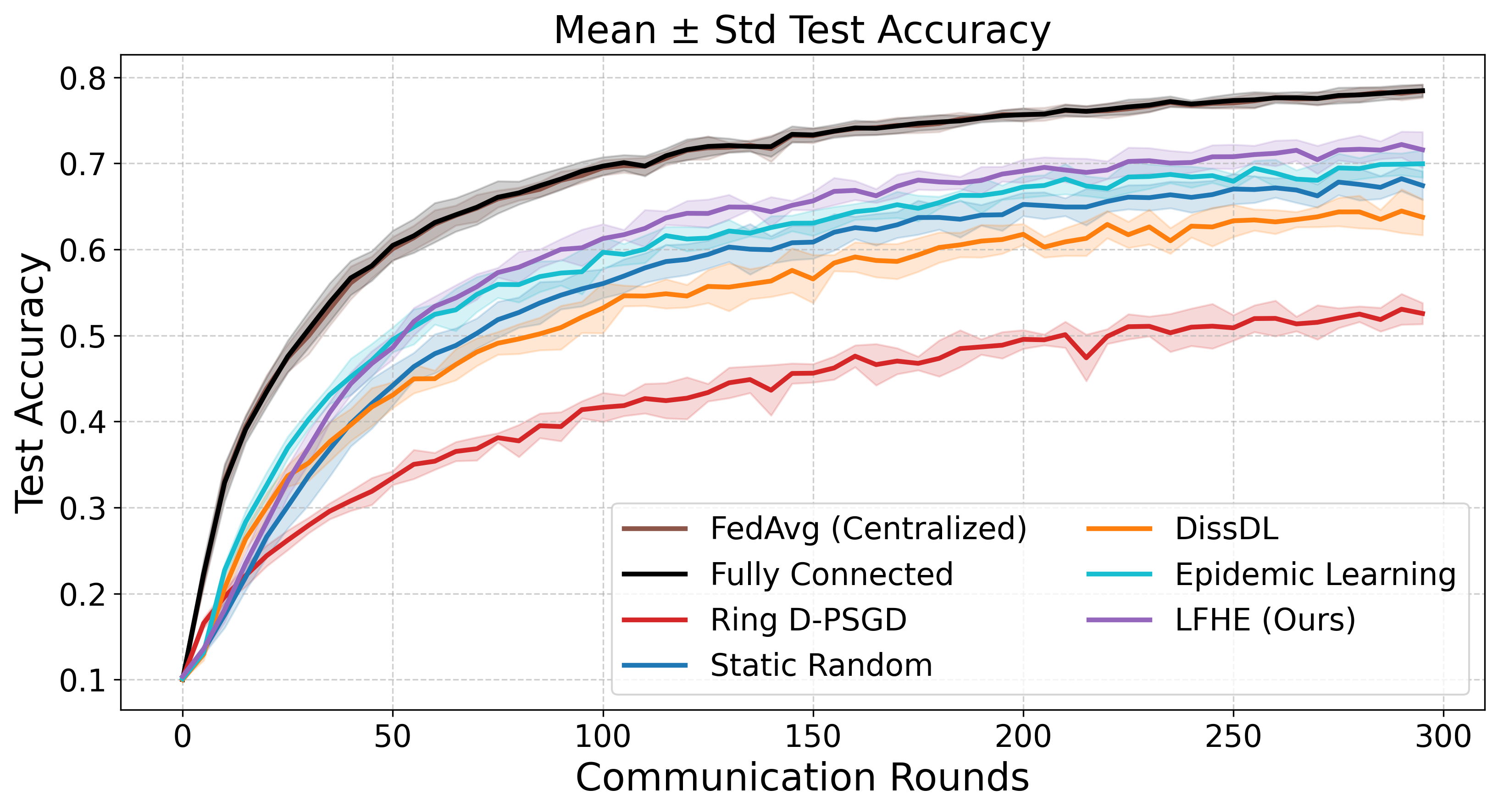}
    \caption*{\small(b) Test accuracy}
\end{minipage}

\caption{\small Main convergence results on CIFAR-10 under non-IID data. LFHE shows favorable long-run performance relative to the reported decentralized baselines; shaded bands summarize run-to-run variability.}
\label{fig:convergence}
\end{figure}

We evaluate LFHE with 30 clients across image, speech, and text tasks. Among the decentralized methods reported in Table~\ref{tab:main_results}, LFHE reaches the earliest or joint-earliest mean target accuracy on all four datasets and obtains the highest mean final accuracy on CIFAR-10, CIFAR-100, and Speech Commands. On Sentiment140, Epidemic Learning has the highest mean final accuracy. Because several differences are comparable to the reported standard deviations, these results support competitive and often favorable performance rather than uniform statistical dominance. Figure~\ref{fig:convergence} illustrates CIFAR-10 dynamics at \(\alpha=0.1\). \looseness-1

\noindent\textbf{Target attainment.}
As shown in Table~\ref{tab:main_results}, LFHE reaches the selected target first among the reported decentralized methods at rounds \textbf{225}, \textbf{125}, and \textbf{170} on CIFAR-10, CIFAR-100, and Sentiment140, respectively, and jointly reaches the Speech Commands target with Epidemic Learning at round \textbf{60}. These measurements are based on communication rounds rather than wall-clock time or bytes-to-target and depend on the selected accuracy thresholds.

\noindent\textbf{Final performance.}
Among the decentralized methods in Table~\ref{tab:main_results}, LFHE achieves the highest mean final accuracy on CIFAR-10, CIFAR-100, and Speech Commands, at \textbf{71.6\%}, \textbf{45.2\%}, and \textbf{78.8\%}, respectively. On Sentiment140, LFHE reaches \textbf{65.3\%}, slightly below Epidemic Learning at \(65.5\%\). The small gaps on several tasks and overlapping run variability indicate strong cross-modal competitiveness rather than consistent superiority over every alternative.

\noindent\textbf{Scalability perspective.}
A preliminary three-seed accuracy study at \(N\in\{10,50,100\}\) is complemented by a discovery-state diagnostic extending to \(N=500\). Across \(N=30\) to \(N=500\), LFHE's measured mean FoF candidate set remains between 9.76 and 11.86 peers per client. This empirical stability is consistent with the \(O(1)\)-per-client discovery-state bound under sparse degree that does not scale with \(N\), showing that LFHE's local candidate space need not grow with population size. This result concerns discovery/control state rather than total communication volume or large-system learning speed. Additional moderate-scale accuracy is provided in Appendix~\ref {appendix:scalability}.

\noindent\textbf{Key insight.}
Taken together, the results separate the effect of localized candidate exposure from that of structural scoring. Random-FoF reaches accuracy close to LFHE in the matched regime, indicating that FoF exposure itself contributes substantially to the observed learning performance. However, degree-preserving random swaps and the graph-only control show that topology churn or high algebraic connectivity alone do not reproduce LFHE's behavior, while the structural score produces substantially different realized communication graphs. Together with the Dirichlet-energy interpretation, these controls support the structural term as a principled state-dependent topology-selection signal whose clearest empirical effect is on communication structure rather than a universal accuracy advantage. Full matched-control trajectories are provided in Appendix~\ref{app:bridge},
with computational and control-overhead measurements in Appendix~\ref{app:cost}.

\begin{table*}[t]
\centering
\scriptsize
\caption{\small Convergence speed and final accuracy across datasets over five random seeds. \textbf{R@T} denotes the communication round at which the mean accuracy across seeds first reaches the target specified in each dataset column, and ``--'' indicates that the target was not reached within the training budget. Final accuracy is reported as mean$\pm$std. FedAvg and Fully Connected are included as reference points. Among decentralized methods, the best result is shown in bold and the second-best in underline.}
\label{tab:main_results}

\scalebox{0.97}{
\begin{tabular}{lcccccccc}
\toprule
\multirow{2}{*}{Method}
& \multicolumn{2}{c}{CIFAR-10 ($\geq$70\%)}
& \multicolumn{2}{c}{CIFAR-100 ($\geq$35\%)}
& \multicolumn{2}{c}{Speech Commands ($\geq$70\%)}
& \multicolumn{2}{c}{Sentiment140 ($\geq$65\%)} \\
\cmidrule(lr){2-3}
\cmidrule(lr){4-5}
\cmidrule(lr){6-7}
\cmidrule(lr){8-9}
& R@T & Final Acc. (\%) & R@T & Final Acc. (\%) & R@T & Final Acc. (\%) & R@T & Final Acc. (\%) \\
\midrule

FedAvg
& 115 & 78.40$\pm$0.80
& 85 & 51.90$\pm$0.30
& 35 & 86.80$\pm$0.50
& 40 & 70.60$\pm$2.70 \\

Fully Connected
& 105 & 78.40$\pm$0.70
& 85 & 51.80$\pm$0.50
& 35 & 86.40$\pm$1.40
& 40 & 70.60$\pm$2.60 \\

\midrule

Ring
& -- & 52.60$\pm$1.20
& -- & 27.00$\pm$0.80
& -- & 53.00$\pm$3.10
& -- & 58.00$\pm$2.70 \\

Static Random
& -- & 67.40$\pm$1.70
& 160 & 41.70$\pm$1.20
& 75 & 74.20$\pm$3.50
& -- & 63.50$\pm$2.30 \\

Epidemic Learning
& -- & \underline{70.00$\pm$1.60}
& \underline{130} & \underline{44.10$\pm$0.70}
& \textbf{60} & \underline{78.10$\pm$3.30}
& \underline{190} & \textbf{65.50$\pm$2.30} \\

DissDL (legacy impl.)
& -- & 63.70$\pm$2.10
& 185 & 40.20$\pm$0.30
& 80 & 74.60$\pm$3.70
& 230 & 64.20$\pm$3.80 \\

\midrule

\textbf{LFHE (Ours)}
& \textbf{225} & \textbf{71.60$\pm$2.10}
& \textbf{125} & \textbf{45.20$\pm$0.60}
& \textbf{60} & \textbf{78.80$\pm$2.30}
& \textbf{170} & \underline{65.30$\pm$1.70} \\

\bottomrule
\end{tabular}
}
\end{table*}

\subsection{Topology Evolution Behavior}

\begin{figure*}[t]
\centering

\begin{minipage}[t]{0.48\textwidth}
    \centering
    \includegraphics[width=0.95\linewidth]{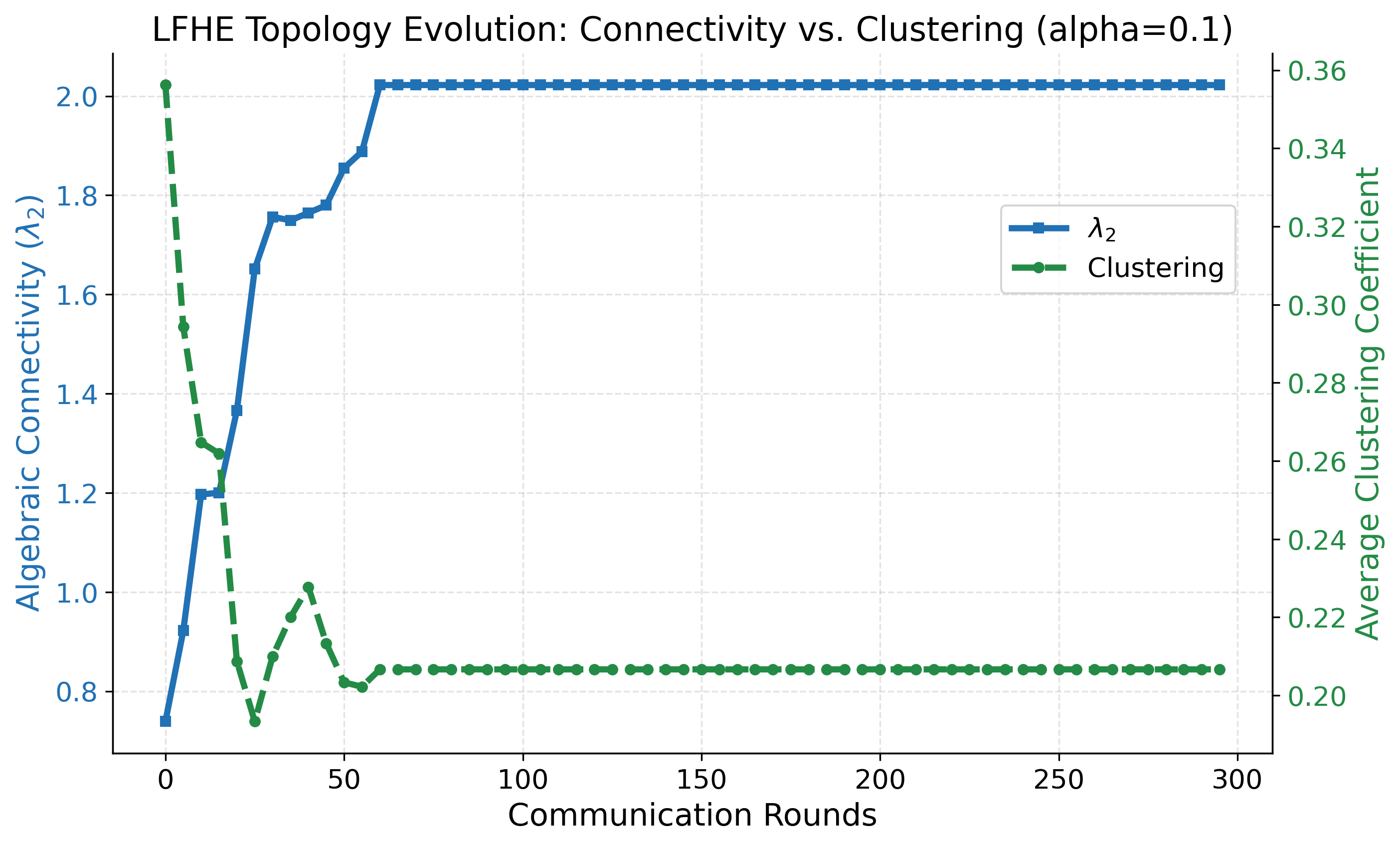}
    \caption*{(a) Evolution of $\lambda_2$ and clustering}
\end{minipage}
\hfill
\begin{minipage}[t]{0.48\textwidth}
    \centering
    \includegraphics[width=0.95\linewidth]{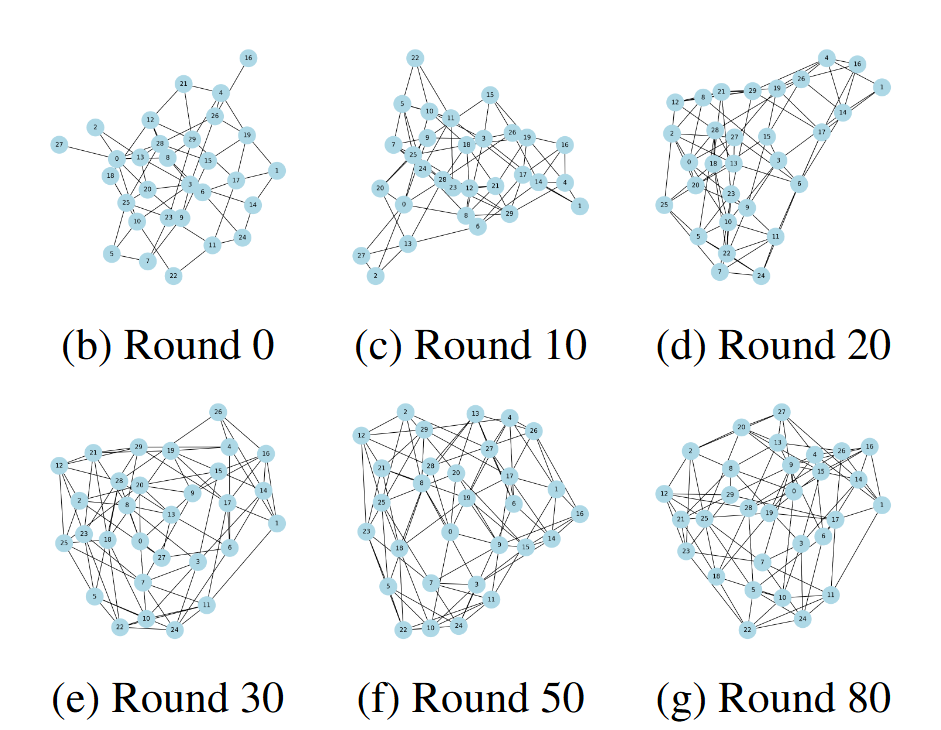}
    \caption*{(b) Topology snapshots}
\end{minipage}

\caption{\small Topology evolution of LFHE on CIFAR-10 under non-IID data ($\alpha=0.1$). Left: algebraic connectivity $\lambda_2$ and clustering coefficient. Right: representative graph snapshots at different training rounds.}
\label{fig:topology_evolution_combined}
\end{figure*}

To examine how LFHE reshapes the communication graph, we analyze global topology statistics and representative graph snapshots under the highly non-IID CIFAR-10 setting ($\alpha=0.1$). Figure~\ref{fig:topology_evolution_combined}(a) tracks the evolution of algebraic connectivity and clustering, while Figure~\ref{fig:topology_evolution_combined}(b) shows representative graph snapshots over training. Additional consensus and client-level variance diagnostics are provided in
Appendix~\ref{appendix:consensus_stability}. \looseness-1

\noindent\textbf{Early improvement in global connectivity.}
As shown in Figure~\ref{fig:topology_evolution_combined}(a)--(b), the offline diagnostic \(\lambda_2\) rises early and then stabilizes, showing that the realized graphs become more algebraically connected. This complements the state-dependent interpretation developed in Section~\ref{sec:consensus_theory}: LFHE scores the current representation field locally, while \(\lambda_2\) independently measures the resulting graph's worst-case spectral connectivity.

\noindent\textbf{Reduced local redundancy.}
The clustering coefficient decreases early and then stabilizes at a lower level. Together with the increase in \(\lambda_2\), this indicates that LFHE restructures local connectivity rather than simply forming denser neighborhoods. The resulting graphs remain sparse while exhibiting communication patterns compatible with broader information propagation across the network.

\noindent\textbf{Exploration-to-refinement dynamics.}
The trajectories and snapshots are consistent with early bridge-like exploration followed by less frequent rewiring as the annealing schedule shifts weight toward the structural term. By rounds 10--20, new cross-region edges coincide with the sharp rise in \(\lambda_2\); later graphs remain sparse and show no dominant hub. These observations are consistent with the intended exploration-to-refinement behavior, while the component diagnostics below separately examine the roles of the individual objective terms. \looseness-1

\subsection{Component and Mechanism Diagnostics}

The original removal ablation on CIFAR-10 with \(\alpha=0.3\) shows that full LFHE reaches the target first and obtains the highest final mean among the reported removal variants. A stricter matched-score diagnostic at \(\alpha=0.1\), however, shows that Structural-only slightly exceeds full LFHE: final \(\lambda_2\) is \(2.306\) versus \(2.247\), and final accuracy is \(75.08\%\) versus \(74.82\%\). Exploration-only rewires more frequently (\(3.406\) versus \(1.333\) accepted rewires per update for LFHE) while producing weaker connectivity and accuracy. These results identify the structural term as the principal empirical topology-selection signal in this regime, with exploration and degree control acting as regime-dependent search regulators rather than universally necessary components.

Additional matched controls reinforce this interpretation without reducing topology quality to a single graph statistic. Random-FoF reaches \(74.62\pm2.25\%\) final accuracy across three seeds, close to LFHE, but produces lower final algebraic connectivity (\(1.696\) versus \(2.247\)). A degree-preserving random-swap control performs many topology changes but reaches only \(66.24\%\) accuracy, while a representation-free graph-only control reaches a higher final \(\lambda_2=2.403\) but only \(73.52\%\) accuracy. Together, these controls show that neither topology churn nor maximizing spectral connectivity alone explains LFHE's behavior, and they support the role of state-dependent structural scoring. Full removal ablations are provided in Appendix~\ref{app:full_ablation},
and matched mechanism controls and trajectories in Appendix~\ref{app:bridge}.

\section{Discussion}

LFHE's structural signal is not merely a heuristic proxy for algebraic connectivity. For the current representation field, \(C_i\) is exactly a node-local contribution to graph Dirichlet energy, whose value gives the instantaneous dissipation rate of centered disagreement under standard linear consensus. LFHE therefore performs state-aware local topology search rather than approximating a global spectral objective.

The experiments support this distinction. Structural-only slightly exceeds full LFHE in the matched-score diagnostic, while Random-FoF, graph-only, and random-swap controls show that neither FoF exposure, high \(\lambda_2\), nor topology churn alone explains the observed behavior. The structural term therefore acts as the principal topology-selection signal in the tested regime, while exploration and degree control regulate how that signal is searched and constrained.

A separate three-seed comparison with Morph
(Appendix~\ref{appendix:morph_comparison}) further illustrates this trade-off. Morph achieves stronger predictive performance, whereas LFHE restricts search to a bounded FoF frontier. Under bounded sparse degree, LFHE maintains \(O(1)\) candidate state per client and \(O(N)\) system-wide, emphasizing discovery locality rather than predictive superiority or lower total communication.

The reference implementation serializes topology proposals, so realized graphs
may depend on update order; Appendix~\ref{appendix:order_endpoint} quantifies
this sensitivity. Fully concurrent or asynchronous execution remains outside the evaluated protocol.

\section{Conclusion}

We introduced LFHE, a framework for bounded local topology search in decentralized learning with non-IID data. LFHE evaluates local add-or-swap operations using structural disagreement, exploration, and degree control without online global spectral computation. Its structural scores exactly decompose representation Dirichlet energy, which under standard linear consensus determines instantaneous disagreement dissipation.

Across the reported baselines, LFHE shows favorable target attainment and competitive final accuracy, while matched diagnostics identify the structural term as the main empirical topology-selection signal. LFHE should therefore be viewed as state-aware localized topology search with bounded FoF discovery, rather than as an approximation to global spectral optimization.

\noindent\textbf{Limitations.}
The theoretical result applies to the structural component with fixed representations during each topology decision and under standard linear consensus. The full objective still depends on representation choice, coefficient settings, and the annealing schedule. \(D_{\max}\) is an online edge-acquisition threshold rather than a strict global degree cap, acceptance is initiator-local, and topology updates are serialized. Evaluation remains limited to controlled benchmarks and moderate training populations.

\noindent\textbf{Future work.} Future directions include the analysis of the discrete weighted aggregation operator, the development of learned exploration schedules, the investigation of concurrent topology updates, and large-scale evaluation under communication-normalized settings.

\subsection*{Reproducibility statement}
The main protocol, optimization settings, and comparison scope are summarized in Section~\ref{sec:experimental_setup}. Complete experimental settings and architectures are provided in Appendix~\ref{appendix:exp_details}, full algorithmic details in Appendix~\ref{app:algorithm}, and sensitivity and diagnostic protocols in Appendices~\ref{appendix:sensitivity}--\ref{appendix:consensus_stability}.

\subsection*{AI use statement}
Generative AI tools were used to assist with language editing, restructuring portions of the manuscript, experiment planning, and software/code edits. AI-assisted text, code changes, and experiment configurations were reviewed by the authors, while numerical results were taken from the executed experimental pipelines and checked against saved outputs. The authors take responsibility for the final text, claims, code, and reported artifacts produced with AI assistance.

\bibliography{references}
\bibliographystyle{plainnat}

\clearpage

\appendix

\section{Full Component Ablation}
\label{app:full_ablation}

\subsection{Original removal ablation and topology diagnostics}

To assess the role of each component, we conduct an ablation on CIFAR-10 with \(\alpha=0.3\). Under this specific protocol, full LFHE reaches the target first and has the highest final mean among the removal variants. This experiment tests removal from the full objective; it should be distinguished from the matched-score diagnostic at \(\alpha=0.1\), where structural-only is slightly stronger than full LFHE.

\begin{table}[t]
\centering

\begin{minipage}[t]{0.50\linewidth}
    \centering
    \vspace{0pt}
    \includegraphics[width=\linewidth]{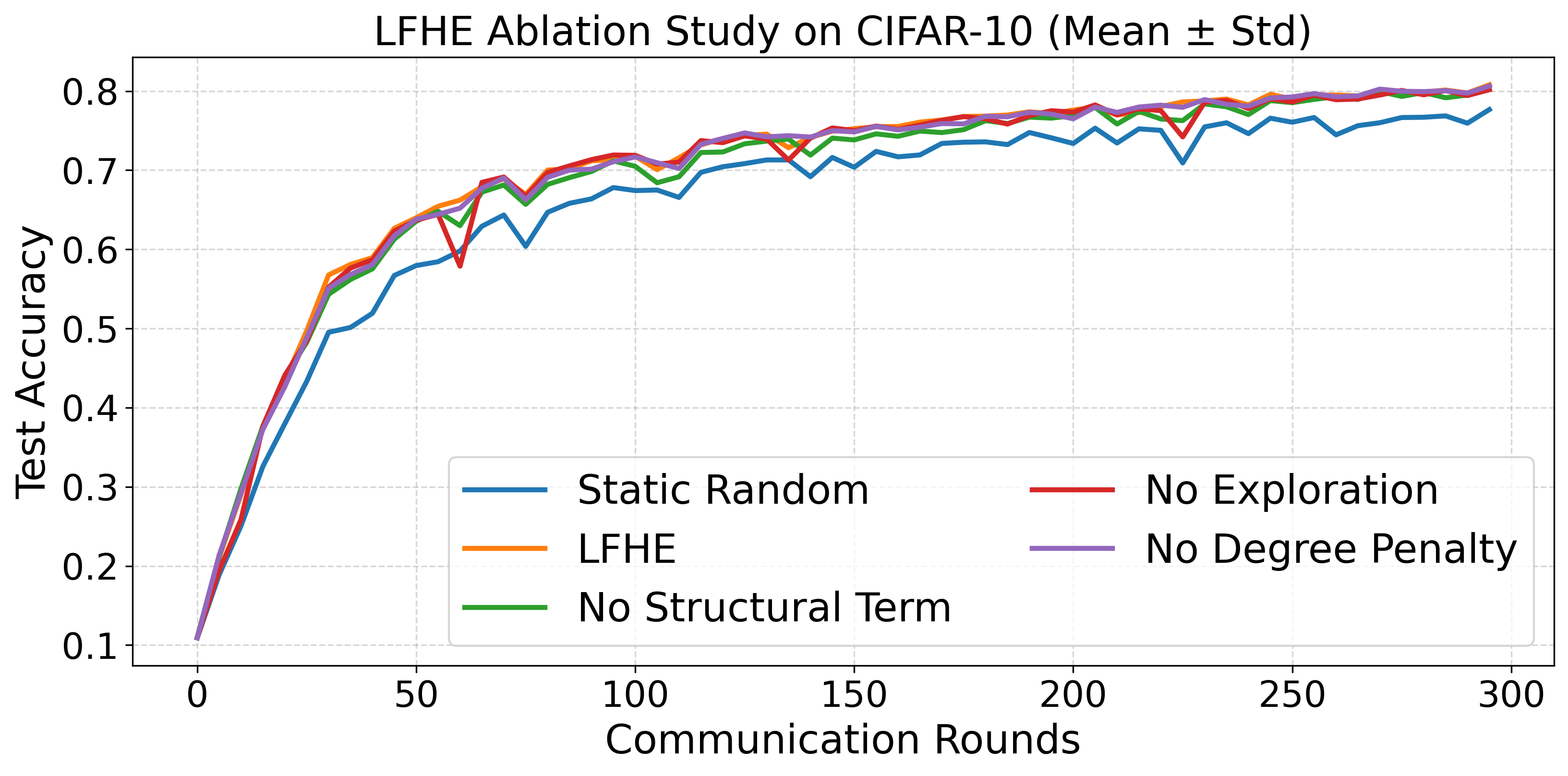}
    \captionof{figure}{\small Ablation study on CIFAR-10 showing test accuracy for LFHE and its ablated variants.}
    \label{fig:ablation_accuracy}
\end{minipage}
\hfill
\begin{minipage}[t]{0.45\linewidth}
    \centering
    \vspace{0pt}
    \small
    \setlength{\tabcolsep}{3pt}
    \begin{tabular}{lcc}
    \toprule
    Method & R@Target & Final Acc. \\
    \midrule
    Static Random   & 85           & 0.7766 \\
    LFHE            & \textbf{55}  & \textbf{0.8079} \\
    w/o Structural  & 65           & 0.8019 \\
    w/o Exploration & 65           & 0.8017 \\
    w/o Degree Reg. & 60           & 0.8063 \\
    \bottomrule
    \end{tabular}
    \vspace{4pt}
    \captionof{table}{\small Ablation results on CIFAR-10 with $\alpha=0.3$. R@Target denotes the round at which accuracy first reaches 0.65, and Final Acc.\ denotes the final test accuracy.}
    \label{tab:ablation}
\end{minipage}

\end{table}

\begin{figure}[h]
\centering
\includegraphics[width=0.4\linewidth]{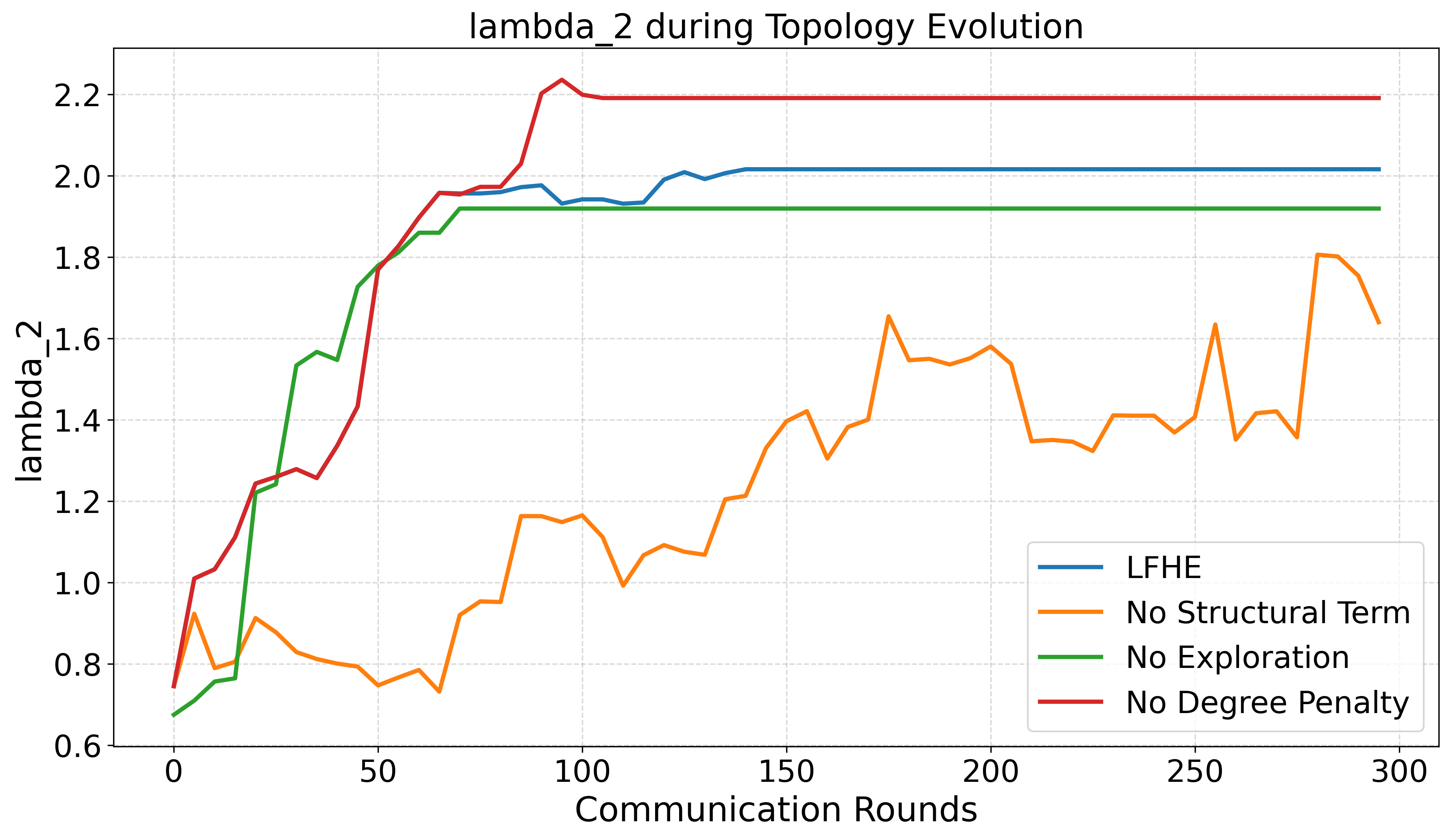}
\includegraphics[width=0.4\linewidth]{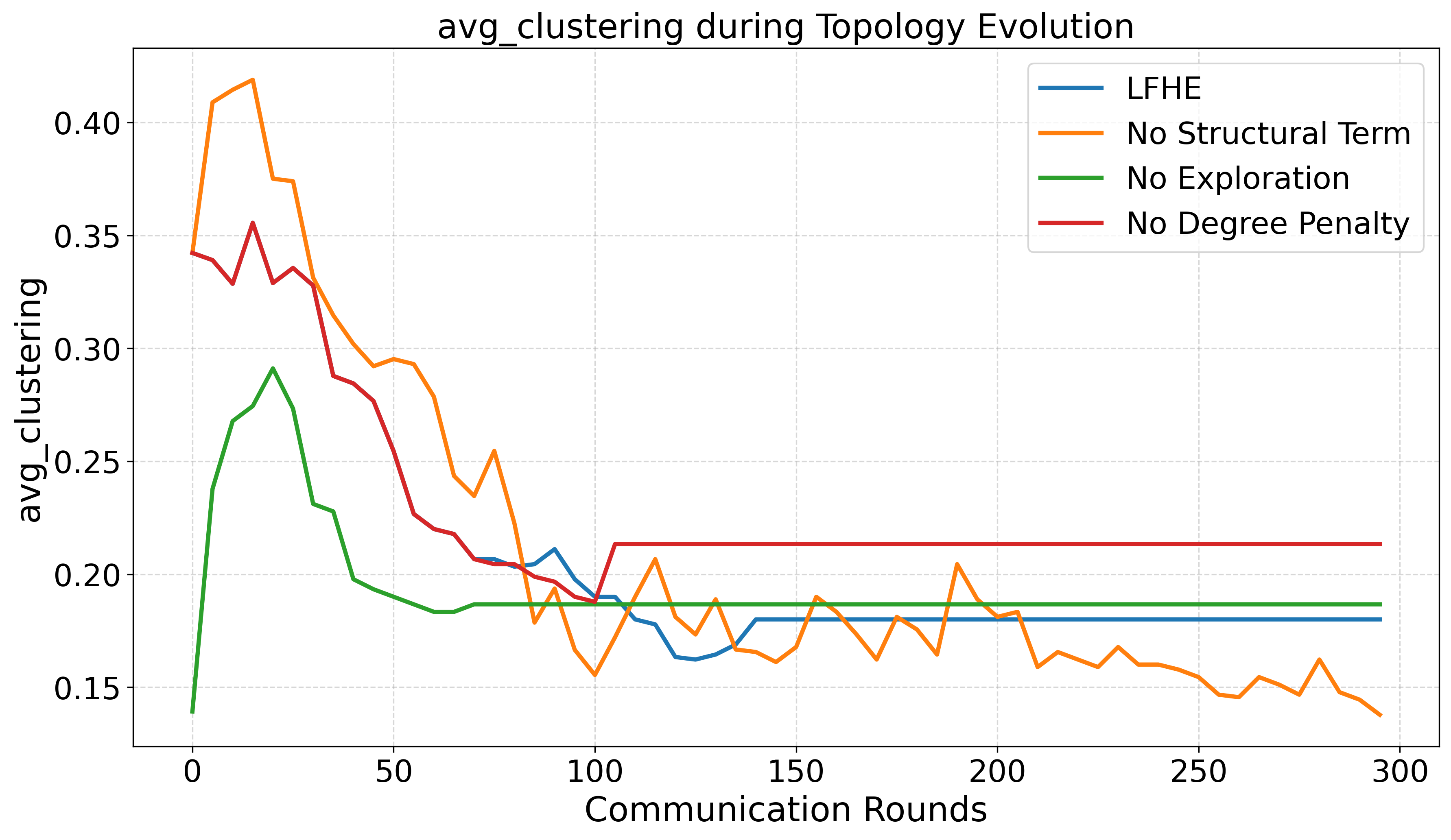}
\caption{\small Topology evolution under ablation. Left: algebraic connectivity (\(\lambda_2\)). Right: clustering coefficient.}
\label{fig:ablation_topology}
\end{figure}

\noindent\textbf{Effect of the local structural term.}
Removing the structural term causes the clearest degradation in topology quality. In Figure~\ref{fig:ablation_topology}, the variant without \(C_i\) attains the weakest and most unstable \(\lambda_2\), together with slower convergence and lower final accuracy in Figure~\ref{fig:ablation_accuracy} and Table~\ref{tab:ablation}. This pattern is consistent with the Dirichlet-energy interpretation in Appendix~\ref{appendix:theory}: removing \(C_i\) removes the objective component that directly scores current representation disagreement across incident communication edges.

\noindent\textbf{Effect of exploration.}
Removing the exploration term produces a mild degradation in this ablation. Without the early exploratory signal, nodes are more likely to remain in locally convenient neighborhoods rather than expose bridge-forming opportunities across heterogeneous regions. The small difference, together with the structural-only result in Appendix~\ref{app:bridge}, indicates that exploration is a regime-dependent search regulator rather than a universally necessary source of final accuracy.

\noindent\textbf{Effect of degree regularization.}
Removing the degree penalty increases \(\lambda_2\) the most, but does not yield the best accuracy. The resulting graph also maintains higher clustering, indicating denser and less controlled connectivity. This indicates that stronger connectivity alone is insufficient; without degree control, the topology becomes denser and less communication-constrained, while the additional connectivity does not translate into improved learning performance~\cite{vogels2022role}.

\noindent\textbf{Topology dynamics and insight.}
Figure~\ref{fig:ablation_topology} identifies the structural term as the clearest source of the measured topology change. Exploration modestly affects search in this regime, while removing degree regularization yields denser connectivity without a corresponding accuracy gain. Importantly, the matched \(\alpha=0.1\) diagnostic reports final \(\lambda_2=2.306\) and accuracy \(0.7508\) for structural-only, compared with \(2.247\) and \(0.7482\) for full LFHE. We therefore do not claim that all three terms are jointly optimal in every regime; full LFHE is the reference composite rule, whereas structural-only can be stronger under the matched-score setting. \looseness-1

\section{Related Work Comparison Table}
\label{app:rw_table}

\begin{table}[h]
\centering
\caption{Comparison of representative communication-structure approaches. The last column refers to online information requirements rather than concurrent execution semantics. \(\checkmark^{*}\) denotes a state-dependent Dirichlet structural signal rather than direct optimization of \(\lambda_2\).}
\small
\begin{tabular}{lcccc}
\toprule
Method & Connectivity-aware & Similarity-aware & Ego-graph local & No global graph statistics \\
\midrule
Spectral methods
& \checkmark & $\times$ & $\times$ & $\times$ \\
D-PSGD / Gossip
& $\times$ & $\times$ & $\times$ & \checkmark \\
D-Cliques / STL-FW
& \checkmark & \checkmark & $\times$ & $\times$ \\
Epidemic Learning
& $\times$ & $\times$ & \checkmark & \checkmark \\
DissDL (legacy impl.)
& $\times$ & \checkmark & \checkmark & \checkmark \\
Morph
& $\times$ & \checkmark & $\times$ & \checkmark \\
FedAMP / Ditto
& $\times$ & \checkmark & $\times$ & $\times$ \\
\midrule
\textbf{LFHE}
& \checkmark$^{*}$ & \checkmark & \checkmark & \checkmark \\
\bottomrule
\end{tabular}
\end{table}

\section{Graph Laplacian, Dirichlet Energy, and Algebraic Connectivity}
\label{app:laplacian}

For an undirected graph \(G=(V,E)\) with \(|V|=n\), let \(A\) be its adjacency matrix, \(D=\mathrm{diag}(d_1,\ldots,d_n)\) its degree matrix, and \(L=D-A\) its combinatorial Laplacian. For a scalar graph signal \(x\in\mathbb{R}^n\),
\begin{equation}
x^\top Lx
=
\frac12\sum_{i=1}^{n}\sum_{j=1}^{n}A_{ij}(x_i-x_j)^2
=
\sum_{(i,j)\in E}(x_i-x_j)^2.
\label{eq:lap_quad_app}
\end{equation}
For vector-valued representations \(R=[r_1^\top,\ldots,r_n^\top]^\top\in\mathbb{R}^{n\times d}\), summing over dimensions gives
\begin{equation}
\mathcal{E}_G(R)
:=
\mathrm{Tr}(R^\top LR)
=
\sum_{(i,j)\in E}\|r_i-r_j\|_2^2,
\label{eq:dirichlet_app}
\end{equation}
the graph Dirichlet energy of the current representation field.

The eigenvalues satisfy \(0=\lambda_1\le\lambda_2\le\cdots\le\lambda_n\). By the Courant--Fischer theorem,
\begin{equation}
\lambda_2(L)
=
\min_{\substack{x\neq0\\x\perp\mathbf1}}
\frac{x^\top Lx}{x^\top x}.
\label{eq:rayleigh_app}
\end{equation}
Thus \(\lambda_2\) is not the energy of the current model state; it is the smallest normalized Dirichlet energy over all non-consensus directions, explaining its role as a topology-only worst-case connectivity and consensus quantity~\cite{olfati2004consensus,olfati2007consensus}.

\section{Structural Signal as Local Consensus Dissipation}
\label{appendix:theory}

LFHE uses
\begin{equation}
C_i(R,G)=\sum_{j\in\mathcal{N}_i(G)}\|r_i-r_j\|_2^2.
\label{eq:ci_app}
\end{equation}

\noindent\textbf{Proposition 1 (Local Dirichlet decomposition).}
For any undirected \(G\) and fixed \(R\),
\begin{equation}
\mathcal{E}_G(R)=\frac12\sum_{i=1}^{n}C_i(R,G).
\label{eq:ci_decomp}
\end{equation}

\noindent\emph{Proof.}
Summing \(C_i\) counts every undirected edge disagreement once from each endpoint:
\[
\sum_i C_i
=
2\sum_{(i,j)\in E}\|r_i-r_j\|_2^2
=
2\,\mathrm{Tr}(R^\top LR).
\]
Dividing by two proves the result. \hfill\(\square\)

\noindent\textbf{Proposition 2 (Consensus-dissipation identity).}
Consider standard continuous-time linear consensus
\begin{equation}
\dot R=-LR.
\end{equation}
Let \(P=I-\frac1n\mathbf1\mathbf1^\top\) and \(V(R)=\frac12\|PR\|_F^2\). Then
\begin{equation}
\dot V(R)=-\mathcal{E}_G(R)=-\frac12\sum_i C_i(R,G).
\label{eq:consensus_dissipation_app}
\end{equation}

\noindent\emph{Proof.}
Since \(L\mathbf1=0\), \(PL=LP=L\), and therefore
\[
\dot V
=
\mathrm{Tr}\!\left((PR)^\top P\dot R\right)
=
-\mathrm{Tr}(R^\top LR)
=
-\mathcal{E}_G(R).
\]
Proposition~1 gives the second equality. \hfill\(\square\)

Thus \(C_i\) is client \(i\)'s local contribution to the instantaneous dissipation of the currently observed representation disagreement. LFHE can therefore allocate limited communication using current model state without reconstructing a global spectral objective.

\noindent\textbf{Corollary (One-edge swap).}
Let \(G'\) replace \((i,j)\) in \(G\) by \((i,k)\), with \(R\) fixed during the topology decision. Then
\begin{equation}
\mathcal{E}_{G'}(R)-\mathcal{E}_G(R)
=
\|r_i-r_k\|_2^2-\|r_i-r_j\|_2^2.
\label{eq:swap_energy}
\end{equation}
A positive structural change therefore strictly increases the instantaneous dissipation rate in Proposition~2. This statement concerns the structural component; the full objective may trade structural change against exploration or degree terms.

Finally, because every column of \(PR\) is orthogonal to \(\mathbf1\), Eq.~\ref{eq:rayleigh_app} gives
\begin{equation}
\mathcal{E}_G(R)
\ge
\lambda_2(L)\|PR\|_F^2
=
2\lambda_2(L)V(R),
\end{equation}
hence \(\dot V\le-2\lambda_2(L)V\). Algebraic connectivity is therefore a worst-case graph-level lower bound on dissipation, whereas \(C_i\) evaluates the representation field currently present during training. LFHE does not approximate \(\lambda_2\); it performs state-aware local structural search within the same Dirichlet-energy framework.

\noindent\textbf{Relation to the implemented averaging rule.}
For fixed \(G\), the degree-aware aggregator is \(W=I-L_a\), with edge weights \(a_{ij}=1/(1+\max(d_i,d_j))\). Its weighted Dirichlet energy is
\[
\mathcal{E}_a(R)=\sum_{(i,j)\in E}a_{ij}\|r_i-r_j\|_2^2.
\]
If \(\bar d=\max_i d_i\), then
\[
\frac{\mathcal{E}_G(R)}{1+\bar d}
\le \mathcal{E}_a(R)
\le \frac12\mathcal{E}_G(R).
\]
Thus bounded degree links LFHE's structural score to the energy induced by the implemented aggregation weights, up to degree-controlled factors.

The experiments use discrete degree-aware averaging rather than the continuous-time model above. The propositions establish the structural score's consensus-theoretic meaning, while the matched FoF controls test its relevance under the implemented learning dynamics. There, Structural-only and LFHE produce the largest measured \(\lambda_2\) gains and lowest model divergence among matched score variants.

\newpage

\section{Full Algorithm Description}
\label{app:algorithm}

In this appendix, we provide the procedural description of the decentralized learning framework with LFHE topology adaptation. Local training and aggregation are synchronous. Within a topology-update call, client proposals are evaluated and committed sequentially on a copied graph, matching the reference implementation used in the experiments.

\subsection{Decentralized Federated Learning Procedure}

\begin{algorithm}[H]
\caption{Decentralized Federated Learning with LFHE Topology Updates}
\label{alg:DFL_appendix}
\begin{algorithmic}[1]
\Require local batch size $B$, local epochs $E$, task-specific local optimizer $\mathcal{O}$, total rounds $T$, topology update interval $K$
\Ensure final local models $\{w_i^T\}_{i \in V}$

\State Initialize a connected communication graph $G^0=(V,E^0)$
\State Initialize local models $\{w_i^0\}_{i \in V}$

\For{each round $t = 0,1,\dots,T-1$}

    \Statex \textit{// Local training}
    \For{each client $i \in V$ in parallel}
        \For{$e = 1,\dots,E$}
            \For{mini-batch $b$ sampled from client $i$}
                \State $w_i \leftarrow \textsc{OptimizerStep}(w_i,b;\mathcal{O})$
            \EndFor
        \EndFor
    \EndFor

    \Statex \textit{// Synchronous decentralized aggregation}
    \For{each client $i \in V$ in parallel}
        \State $N_i \leftarrow \mathcal{N}_i(G^t)$
        \For{each $j \in N_i$}
            \State $a_{ij} \leftarrow \dfrac{1}{1+\max(\deg(i),\deg(j))}$
        \EndFor
        \State $a_{ii} \leftarrow 1 - \sum_{j \in N_i} a_{ij}$
        \State $\tilde{w}_i \leftarrow a_{ii} w_i + \sum_{j \in N_i} a_{ij} w_j$
    \EndFor

    \For{each client $i \in V$}
        \State $w_i \leftarrow \tilde{w}_i$
    \EndFor

    \Statex \textit{// Periodic serialized topology update}
    \If{$t \bmod K = 0$}
        \State $G^{t+1} \leftarrow \textsc{LFHE-Update}(G^t,\{w_i\},t)$
    \Else
        \State $G^{t+1} \leftarrow G^t$
    \EndIf

\EndFor

\State \Return $\{w_i^T\}_{i \in V}$
\end{algorithmic}
\end{algorithm}

\subsection{LFHE Topology Update}
\label{app:lfhe_update}

LFHE performs local graph rewiring using only ego-network information. For each client $i$, a candidate node is discovered through a FoF mechanism. The candidate edge is accepted if it improves the node-level fitness under the degree constraint. When direct addition is not feasible, LFHE attempts an atomic swap by replacing one existing neighbor with the candidate.

\begin{algorithm}[H]
\caption{LFHE Topology Update with Degree Constraint and Atomic Swap}
\label{alg:LFHE_appendix}
\begin{algorithmic}[1]
\Require graph $G=(V,E)$, client models $\{w_i\}$, online degree threshold $D_{\max}$, fitness weights $\omega_1,\omega_2,\omega_3$, round $t$
\Ensure updated graph $G'$

\State $G' \leftarrow G$

\For{each client $i \in V$}

    \State $N_i \leftarrow \mathcal{N}_i(G')$
    \If{$|N_i| = 0$}
        \State \textbf{continue}
    \EndIf

    \Statex \textit{// Friend-of-Friend candidate discovery}
    \State Sample $j \sim \mathrm{Unif}(N_i)$
    \State $\mathcal{C}_i \leftarrow \{k \in \mathcal{N}_j(G') \mid k \neq i,\ (i,k)\notin E(G')\}$
    \If{$\mathcal{C}_i = \emptyset$}
        \State \textbf{continue}
    \EndIf
    \State Sample $k \sim \mathrm{Unif}(\mathcal{C}_i)$

    \State $f_{\mathrm{old}} \leftarrow \textsc{Fitness}(i,G',\{w_i\},t)$

    \If{$\deg(i) < D_{\max}$ and $\deg(k) < D_{\max}$}
        \State Add edge $(i,k)$ to $G'$
        \State $f_{\mathrm{new}} \leftarrow \textsc{Fitness}(i,G',\{w_i\},t)$
        \If{$f_{\mathrm{new}} \le f_{\mathrm{old}}$}
            \State Remove edge $(i,k)$ from $G'$
        \EndIf

    \Else
        \State $f_{\mathrm{best}} \leftarrow -\infty$
        \State $j^\star \leftarrow \text{None}$

        \For{each $j' \in N_i$}
            \If{$\deg(i) \le 1$}
                \State \textbf{continue}
            \EndIf
            \If{$\deg(j') \le 1$}
                \State \textbf{continue}
            \EndIf
            \If{$\deg(k) \ge D_{\max}$}
                \State \textbf{continue}
            \EndIf
            \If{\textbf{not} \textsc{AlternativePath}$(G',i,j')$}
                \State \textbf{continue}
            \EndIf

            \State Temporarily remove $(i,j')$ and add $(i,k)$
            \State $f_{\mathrm{swap}} \leftarrow \textsc{Fitness}(i,G',\{w_i\},t)$

            \If{$f_{\mathrm{swap}} > f_{\mathrm{best}}$}
                \State $f_{\mathrm{best}} \leftarrow f_{\mathrm{swap}}$
                \State $j^\star \leftarrow j'$
            \EndIf

            \State Restore the original edges
        \EndFor

        \If{$f_{\mathrm{best}} > f_{\mathrm{old}}$}
            \State Remove edge $(i,j^\star)$ from $G'$
            \State Add edge $(i,k)$ to $G'$
        \EndIf
    \EndIf

\EndFor

\State \Return $G'$
\end{algorithmic}
\end{algorithm}

\subsection{Fitness Function}
\label{app:fitness}

For each client $i$, LFHE evaluates the quality of its local neighborhood using a node-level fitness function. Let $r_i \in \mathbb{R}^d$ denote the representation vector extracted from client $i$'s local model. The annealing coefficient is defined as
\[
\beta_t = \beta_0 e^{-\kappa t},
\]
where $\beta_0>0$ is the initial annealing weight and $\kappa>0$ is the decay constant.

The first component is a local structural term,
\[
C_i(G) = \sum_{j \in \mathcal{N}_i(G)} \|r_i-r_j\|_2^2,
\]
which measures the incident-edge contribution to representation Dirichlet energy. As shown in Appendix~\ref{appendix:theory}, its network-wide sum has an exact consensus-dissipation interpretation; it is state-dependent and does not require computing a global spectral statistic.

The second component is a neighborhood mismatch term, implemented through cosine similarity to the mean neighbor representation:
\[
\bar{r}_i = \frac{1}{|\mathcal{N}_i(G)|}\sum_{j \in \mathcal{N}_i(G)} r_j,
\]
\[
\mathrm{sim}_i(G) = \cos(r_i,\bar{r}_i).
\]
The corresponding mismatch signal \(1-\mathrm{sim}_i(G)\) acts primarily as an exploration term.

The third component is a degree penalty,
\[
\mathrm{deg\_norm}_i = \frac{\deg(i)}{D_{\max}},
\]
which discourages overly dense local connectivity.

The final fitness used in the implementation is
\[
\textsc{Fitness}(i,G,\{w_i\},t)
=
w_1(1-\beta_t)C_i(G)
+
w_2\beta_t\bigl(1-\mathrm{sim}_i(G)\bigr)
-
w_3 \frac{\deg(i)}{D_{\max}}.
\]

This formulation yields a time-varying trade-off. In early rounds, when $\beta_t$ is relatively large, the mismatch term \(1-\mathrm{sim}_i(G)\) has stronger influence and encourages exploration. In later rounds, as $\beta_t$ decays, the local structural term becomes more influential, promoting topology refinement while still penalizing excessive degree growth.

\subsection{Auxiliary Procedures}
\label{app:auxiliary}

The implementation uses a lightweight local connectivity safeguard before accepting a swap. In particular, the procedure \textsc{AlternativePath}$(G,i,j)$ returns true if clients $i$ and $j$ share at least one common neighbor:
\[
\textsc{AlternativePath}(G,i,j)=
\begin{cases}
\text{true}, & \text{if } \mathcal{N}_i(G)\cap\mathcal{N}_j(G)\neq\emptyset,\\
\text{false}, & \text{otherwise.}
\end{cases}
\]
This is a local redundancy safeguard when an edge is removed. A shared neighbor certifies a two-hop alternative path between the endpoints of the specific edge at the time of the decision, without requiring a global path search. This statement concerns the serialized reference update; stale or conflicting concurrent removals would require additional validation.

\subsection{Implementation Notes}
\label{app:implementation_notes}

In the reference implementation, \textsc{LFHE-Update} is executed on a copy of the current graph and applied sequentially over clients in fixed order. Each accepted operation is committed immediately, so later clients observe earlier changes and the realized topology can depend on processing order. Candidate discovery remains restricted to FoF nodes. If direct addition is infeasible, the initiator evaluates feasible one-edge swaps and commits the swap with the largest improvement in its own fitness. The notation $\{w_i\}$ in Algorithms~\ref{alg:DFL_appendix}--\ref{alg:LFHE_appendix} denotes simulator state compactly; an initiator-local fitness evaluation uses only the representations associated with its current ego-neighborhood and the current FoF candidate.

The affected endpoints do not separately approve the operation, so acceptance is initiator-local rather than bilaterally Pareto-improving. In a controlled update-order diagnostic, changing the client order substantially changed the final edge set and \(\lambda_2\), while the three observed final accuracies stayed within a 1.51-percentage-point range; details are reported in Appendix~\ref{appendix:order_endpoint}. Across the recorded topology audit, none of 2,280 graph states from 38 runs was disconnected. This is empirical evidence about the observed trajectories, not a connectivity theorem for the common-neighbor safeguard.

A concurrent realization could forward FoF candidate identifiers and representation summaries through current neighbors, and could use proposal--validate--commit with endpoint-local topology versions to reject stale or conflicting operations. We present this only as a deployment direction. The experiments in this paper do not implement or validate arbitrary asynchronous transaction semantics.

\section{Additional Experimental Details}
\label{appendix:exp_details}

We provide additional details on datasets, non-IID partitioning, training protocols, model architectures, and implementation hyperparameters used in the experiments.

\paragraph{Datasets.}
\textbf{CIFAR-10} and \textbf{CIFAR-100}~\cite{krizhevsky2009learning} are standard image classification benchmarks derived from the Tiny Images dataset. CIFAR-10 contains 50,000 training images and 10,000 test images across 10 classes, while CIFAR-100 uses the same train--test split structure but has 100 classes. For \textbf{Google Speech Commands}~\cite{warden2018speech}, we use a 10-keyword subset consisting of \textit{yes, no, up, down, left, right, on, off, stop, go}, following the standard training and testing splits. For \textbf{Sentiment140}~\cite{go2009twitter}, we use a binary sentiment classification setting and apply balanced random subsampling before client partitioning, resulting in 12,000 training tweets and 2,000 test tweets.

\paragraph{Non-IID partitioning.}
To simulate heterogeneous client distributions, data are partitioned using class-wise Dirichlet sampling. For each class $c$, we sample a client allocation vector $p_c \sim \mathrm{Dir}(\alpha)$ and distribute samples of that class across clients accordingly. Unless otherwise stated, we use $\alpha=0.1$ as the default highly non-IID setting. The component ablation and moderate-scale diagnostic use $\alpha=0.3$ and are reported separately from the main protocol.

\paragraph{System setting.}
Unless otherwise specified, experiments use \(N=30\) clients and online degree threshold \(D_{\max}=4\). Static Random and LFHE start from the same connected Erd\H{o}s--R\'enyi graph with expected degree four. Because this initializer is not degree-capped, some initial nodes may exceed \(D_{\max}\); the threshold regulates subsequent additions and swaps rather than retroactively enforcing a hard cap. We additionally report a preliminary study at \(N\in\{10,50,100\}\). Ring, Fully Connected, FedAvg, and dynamic baselines retain their respective topology mechanisms.

\paragraph{Training protocol.}
All methods are implemented in PyTorch. CIFAR and Speech Commands experiments use SGD with learning rate 0.05 and batch size 32. Sentiment140 uses Adam with learning rate $10^{-3}$ and batch size 16. Unless otherwise stated, each client performs one local epoch per communication round, except Sentiment140 where each client performs two local epochs. LFHE updates the communication topology every $K=5$ communication rounds. Results are averaged over five random seeds, $\{42,43,44,45,46\}$. Unless otherwise stated, all LFHE experiments use $\omega_1=1.0$, $\omega_2=1.0$, $\omega_3=0.1$, $D_{\max}=4$, and the annealing schedule $\beta_t=\exp(-0.01t)$.

\begin{table}[!htbp]
\centering
\small
\caption{Implementation details and hyperparameters used in the main experiments.}
\label{tab:implementation_details}
\begin{tabular}{p{0.30\linewidth}p{0.62\linewidth}}
\toprule
\textbf{Item} & \textbf{Setting} \\
\midrule
Number of clients & $N=30$ for the main experiments; $N \in \{10,50,100\}$ for scalability experiments. \\
Online degree threshold & \(D_{\max}=4\); the expected-degree initializer is not guaranteed to satisfy a hard cap. \\
Topology update interval & $K=5$ communication rounds. \\
Evaluation interval & Every 5 communication rounds. \\
Random seeds & $\{42,43,44,45,46\}$. \\
Non-IID partitioning & Class-wise Dirichlet partitioning with $\alpha=0.1$ by default; the component ablation and moderate-scale diagnostic use $\alpha=0.3$. \\
Initial sparse graph & Static Random and LFHE use the same connected Erd\H{o}s--R\'enyi graph with expected degree 4 and seed matched to the experimental seed. Ring uses a cycle graph, Fully Connected uses a complete graph, and FedAvg uses centralized aggregation. \\
Aggregation rule & Degree-aware decentralized averaging with $a_{ij}=1/(1+\max(\deg(i),\deg(j)))$ and $a_{ii}=1-\sum_{j\in N_i}a_{ij}$. \\
LFHE weights & $\omega_1=1.0$, $\omega_2=1.0$, $\omega_3=0.1$. \\
Annealing schedule & $\beta_t=\beta_0 e^{-\kappa t}$ with $\beta_0=1.0$ and $\kappa=0.01$. \\
LFHE update rule & Each client samples a FoF candidate, evaluates the local fitness, and accepts the candidate only if the fitness improves. If both endpoints are below the online degree threshold, LFHE adds the edge; otherwise, it evaluates feasible one-edge swaps and commits the best improving swap. \\
CIFAR optimizer & SGD with learning rate 0.05, batch size 32, and one local epoch per communication round. \\
Speech optimizer & SGD with learning rate 0.05, batch size 32, and one local epoch per communication round. \\
Sentiment140 optimizer & Adam with learning rate $10^{-3}$, batch size 16, and two local epochs per communication round. \\
Training rounds & 300 rounds for CIFAR-10, CIFAR-100, and Sentiment140; 100 rounds for Google Speech Commands. \\
CIFAR preprocessing & Random crop with padding 4, random horizontal flip, tensor conversion, and dataset-specific normalization. \\
Speech preprocessing & 10-keyword subset of Speech Commands using 64-bin Mel spectrograms followed by amplitude-to-dB conversion. Variable-length spectrograms are padded within each batch. \\
Sentiment140 preprocessing & Binary labels are mapped from $\{0,4\}$ to $\{0,1\}$, followed by balanced subsampling with 12,000 training tweets and 2,000 test tweets. Text is tokenized with the \texttt{distilbert-base-uncased} tokenizer using maximum length 32. \\
\bottomrule
\end{tabular}
\end{table}

\begin{table}[!htbp]
\centering
\small
\caption{Model architectures and representation extraction rules.}
\label{tab:model_architectures}
\begin{tabular}{p{0.21\linewidth}p{0.51\linewidth}p{0.20\linewidth}}
\toprule
\textbf{Dataset} & \textbf{Model architecture} & \textbf{Representation $r_i$} \\
\midrule
CIFAR-10 &
CNN with two convolutional blocks: Conv(3,32)-BN-ReLU, Conv(32,32)-BN-ReLU, MaxPool; Conv(32,64)-BN-ReLU, Conv(64,64)-BN-ReLU, MaxPool; adaptive average pooling to $4\times4$; MLP classifier $1024 \rightarrow 256 \rightarrow 10$ with dropout 0.3. &
Flattened final classifier-layer weights. \\
\midrule
CIFAR-100 &
Same CNN backbone as CIFAR-10, with the final classifier dimension changed to 100 classes: $1024 \rightarrow 256 \rightarrow 100$. &
Mean-reduced final classifier-layer weights, producing a 100-dimensional representation. \\
\midrule
Speech Commands &
CNN over log-Mel spectrograms: Conv(1,32)-BN-ReLU, Conv(32,32)-BN-ReLU, MaxPool; Conv(32,64)-BN-ReLU, Conv(64,64)-BN-ReLU, MaxPool; adaptive average pooling to $4\times4$; MLP classifier $1024 \rightarrow 256 \rightarrow 10$ with dropout 0.3. &
Flattened final classifier-layer weights. \\
\midrule
Sentiment140 &
Embedding-based text classifier with 128-dimensional token embeddings, attention-mask mean pooling, and MLP classifier $128 \rightarrow 128 \rightarrow 2$ with dropout 0.2. &
Flattened final classifier-layer weights. \\
\bottomrule
\end{tabular}
\end{table}

\paragraph{Compute resources.}
All experiments were implemented in PyTorch 2.2.2+cu118 with Python 3.12.0 and run in a Python virtual environment on a Windows workstation equipped with an NVIDIA GeForce RTX 4060 Laptop GPU with 8GB of GPU memory. The installed NVIDIA driver version was 560.76. PyTorch used CUDA 11.8, while the system driver reported CUDA 12.6 support. The main experiments were averaged over five random seeds, and scalability and overhead analyses used three random seeds where specified. The additional runtime overhead of LFHE topology updates is reported separately in Appendix~\ref{app:cost}.

\section{Additional Experimental Results}
\label{appendix:additional_experiments}

\subsection{Results on Additional Datasets}

\begin{figure}[t]
\centering
\begin{subfigure}{0.49\linewidth}
\centering
\includegraphics[width=\linewidth]{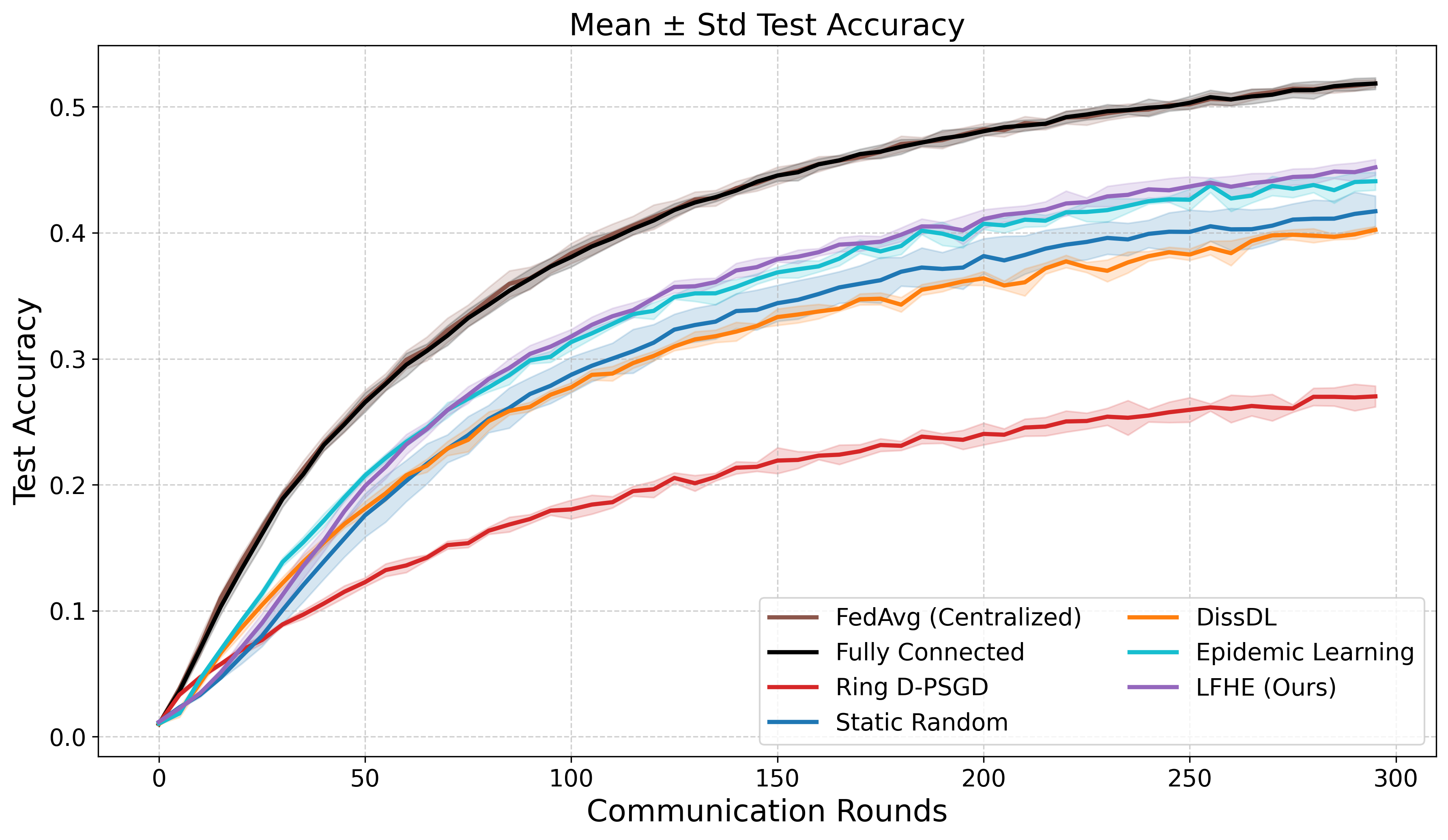}
\caption{CIFAR-100}
\end{subfigure}
\hfill
\begin{subfigure}{0.49\linewidth}
\centering
\includegraphics[width=\linewidth]{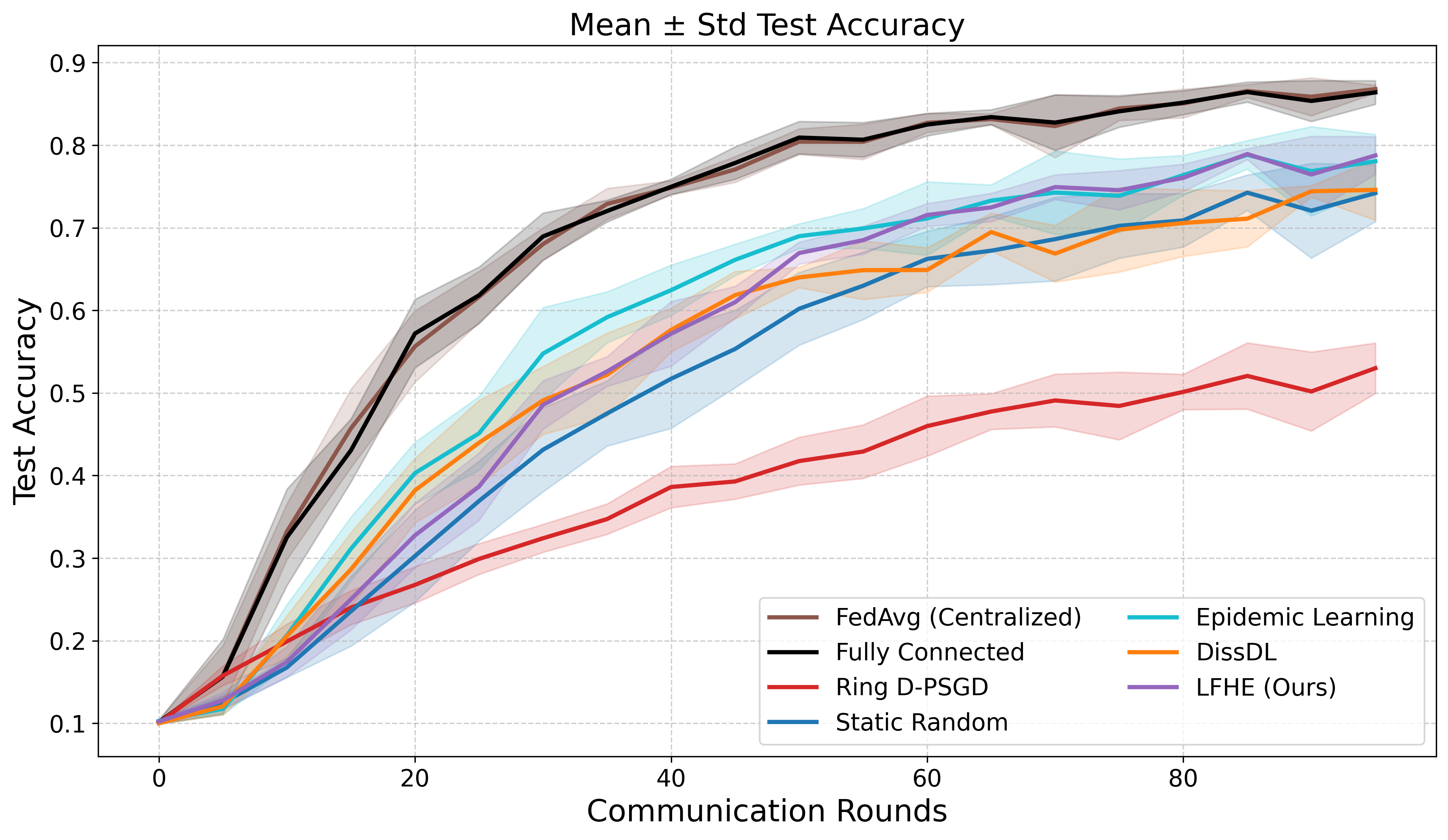}
\caption{Google Speech Commands}
\end{subfigure}

\vspace{0.8em}

\begin{subfigure}{0.49\linewidth}
\centering
\includegraphics[width=\linewidth]{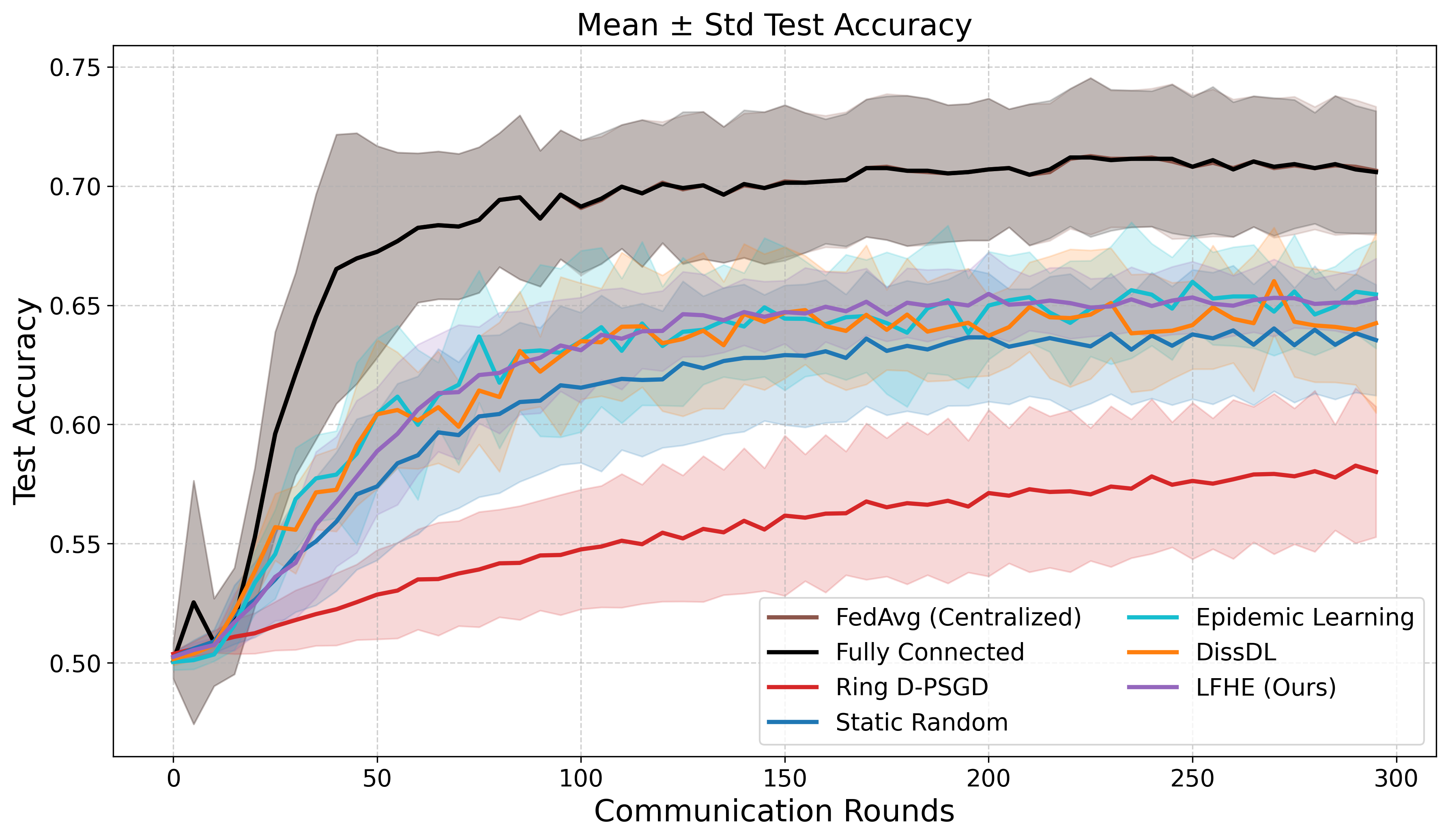}
\caption{Sentiment140}
\end{subfigure}

\caption{\small Test accuracy (mean $\pm$ std) on additional datasets under non-IID data. LFHE is competitive across image, speech, and text tasks; Sentiment140 illustrates that faster target attainment need not imply the best final accuracy.}
\label{fig:additional_datasets_convergence}
\end{figure}

To assess behavior beyond CIFAR-10, we evaluate LFHE on \textbf{CIFAR-100}, \textbf{Google Speech Commands}, and \textbf{Sentiment140}. Figure~\ref{fig:additional_datasets_convergence} reports the mean test-accuracy trajectories. LFHE remains competitive across all three modalities. On CIFAR-100 it reaches the 0.35 target at round 125 and finishes at 45.19\%, ahead of the reported decentralized baselines. On Speech Commands it reaches 0.70 at round 60, jointly earliest with Epidemic Learning, and finishes at 78.76\%. On Sentiment140, LFHE reaches 0.65 at round 170 versus 190 for Epidemic Learning, but Epidemic Learning finishes slightly higher (65.45\% versus 65.29\%). The text setting is therefore a convergence advantage rather than a final-accuracy win.

\subsection{Direct Comparison with Morph}
\label{appendix:morph_comparison}

Morph~\cite{cox2026morph} is the closest recent adaptive comparison, but it does not share LFHE's discovery protocol. We therefore report it as a separate systems comparison rather than fold it into the matched FoF controls. Under a shared three-seed learning protocol, Morph achieves stronger predictive performance than LFHE.

\begin{table}[h]
\centering
\small
\caption{Three-seed LFHE--Morph comparison under the shared diagnostic protocol. nAUC denotes normalized accuracy area under the learning curve.}
\label{tab:morph_comparison}
\begin{tabular}{lcc}
\toprule
Method & Final accuracy (\%) & nAUC (\%) \\
\midrule
LFHE & 71.37$\pm$2.19 & 60.57$\pm$1.52 \\
Morph & \textbf{72.25$\pm$1.92} & \textbf{63.17$\pm$1.23} \\
\bottomrule
\end{tabular}
\end{table}

Morph also reaches 65\% accuracy 16.7 rounds earlier on average. We therefore do not claim predictive superiority over Morph. The distinction lies in peer-discovery scope. From \(N=30\) to \(N=500\), LFHE's measured mean FoF candidate set remains 9.76--11.86 peers per client, whereas Morph's discovered-peer state grows from 29 to 499 peers per client. The corresponding system-wide discovery entries grow from 480 to 8,000 for LFHE and from 3,480 to 998,000 for Morph. This comparison concerns discovery/control state rather than total communication traffic: broader discovery can improve peer selection, and the present evidence does not show that LFHE uses less total communication overall.

\subsection{Update-Order and Endpoint Effects}
\label{appendix:order_endpoint}

The reference simulator serializes topology changes, so we test fixed, reverse, and independently randomized per-update client orders on a controlled CIFAR-10 seed while holding the data partition, initialization, training configuration, candidate generation, acceptance rule, and topology-update frequency fixed.

\begin{table}[h]
\centering
\small
\caption{Single-seed update-order diagnostic. Edge Jaccard is measured against the final graph from fixed order.}
\label{tab:order_sensitivity}
\begin{tabular}{lccccc}
\toprule
Order & Final acc. (\%) & nAUC (\%) & R@65 & Final $\lambda_2$ & Edge Jaccard \\
\midrule
Fixed & 69.03 & 58.82 & 160 & 0.936 & 1.000 \\
Reverse & 70.54 & 60.29 & 145 & 0.685 & 0.291 \\
Random/update & 69.61 & 59.24 & 155 & 0.807 & 0.271 \\
\bottomrule
\end{tabular}
\end{table}

All three final graphs remain connected. The exact topology is substantially order-dependent, whereas the observed final accuracies remain within 1.51 percentage points in this diagnostic. Acceptance is also one-sided: at least one affected non-initiator endpoint has a negative local-fitness change in 33.3\%, 31.5\%, and 41.5\% of accepted operations for fixed, reverse, and randomized processing, respectively; the added-endpoint rates are 33.3\%, 29.6\%, and 39.0\%. Thus, positive initiator gain should not be interpreted as bilateral utility improvement.

\subsection{Sensitivity and Communication Trade-off}
\label{appendix:sensitivity}

We use a separate three-seed CIFAR-10 diagnostic to vary one LFHE design factor at a time while keeping the remaining protocol fixed. Varying either the structural coefficient \(\omega_1\) or exploration coefficient \(\omega_2\) from \(0.5\times\) to \(2\times\) its default keeps final accuracy within 78.01--78.32\%, compared with 78.11$\pm$1.84\% for the default, and keeps R@65 within 71.7--73.3 rounds. Halving the degree coefficient from 0.1 to 0.05 gives 77.64$\pm$1.46\% and R@65 of 71.7. A slower exponential decay gives 78.19\% mean accuracy, 67.21\% nAUC, and R@65 of 71.7, compared with 78.11\%, 67.23\%, and 73.3 for the reference schedule. These measurements support a locally stable region around the chosen coefficients; they do not establish cross-dataset hyperparameter invariance.

The degree threshold exposes a clearer resource trade-off. Increasing the online threshold from 4 to 6 raises three-seed mean accuracy from 78.11\% to 79.70$\pm$1.28\%, nAUC from 67.23\% to 69.38\%, and final \(\lambda_2\) from 1.086 to 1.942, while reducing R@65 from 73.3 to 63.3 rounds. In the seed-42 communication audit, model transmissions increase from 37,810 to 52,540, a 39\% increase. Thus, additional degree capacity improves this diagnostic's learning metrics at a material communication cost.

\subsection{Moderate-Scale Accuracy Study}
\label{appendix:scalability}

\begin{figure}[h]
\centering
\includegraphics[width=0.75\linewidth]{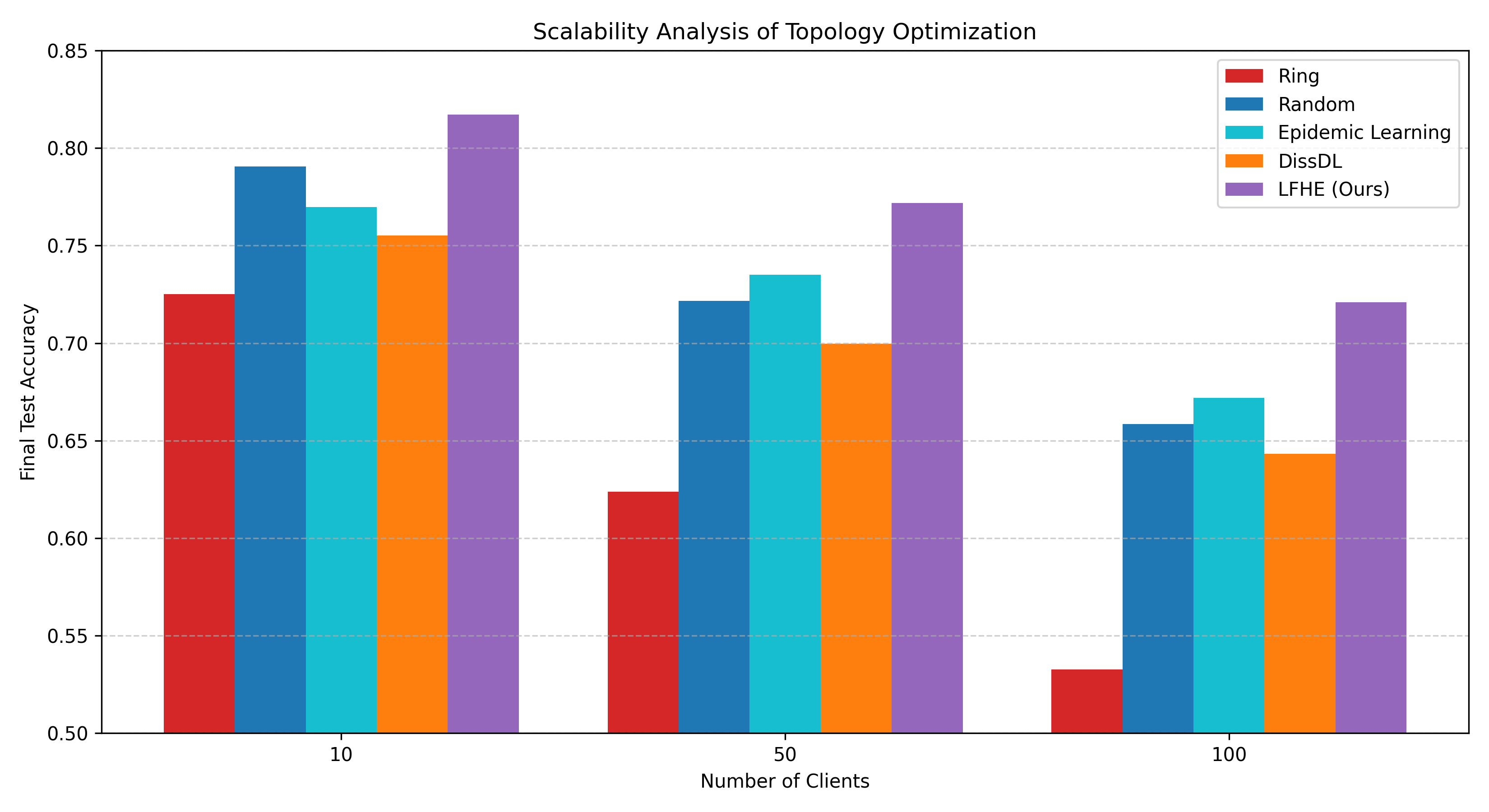}
\caption{Moderate-scale CIFAR-10 diagnostic under $\alpha=0.3$ with $N \in \{10,50,100\}$. Bars report final test accuracy averaged over three seeds.}
\label{fig:scalability}
\end{figure}

We report a preliminary three-seed CIFAR-10 study with \(\alpha=0.3\) and \(N\in\{10,50,100\}\). Figure~\ref{fig:scalability} reports final test accuracy. LFHE has the highest mean final accuracy among the plotted methods at these three scales. Because the experiment stops at \(N=100\), uses only three seeds, and does not normalize the comparison by total communication, it is evidence of moderate-scale behavior rather than a large-system scaling law.

\subsection{Deployment-Stress Diagnostic}
\label{appendix:stress}

We test one compounded communication-disruption condition on seed 42 without retuning LFHE: 80\% client participation, 10\% link-failure probability, and buffered 0--1-round message delay. The graph remains connected throughout the run and the minimum observed \(\lambda_2\) is 0.670, despite 2,472 dropped model messages, 147 dropped control messages, and 636 delayed or stale control messages. Final accuracy decreases from 79.53\% in the clean diagnostic to 60.18\% under stress.

This result separates topological operation from predictive robustness. The observed topology remains connected, but learning performance degrades substantially. Because this is a single-seed compounded stress test, it is preliminary evidence only and does not establish robustness to arbitrary asynchrony, long delays, or real distributed failures.

\subsection{Additional Consensus and Client-Variance Diagnostics}
\label{appendix:consensus_stability}

\begin{figure}[t]
\centering
\begin{subfigure}{0.48\linewidth}
    \centering
    \includegraphics[width=\linewidth]{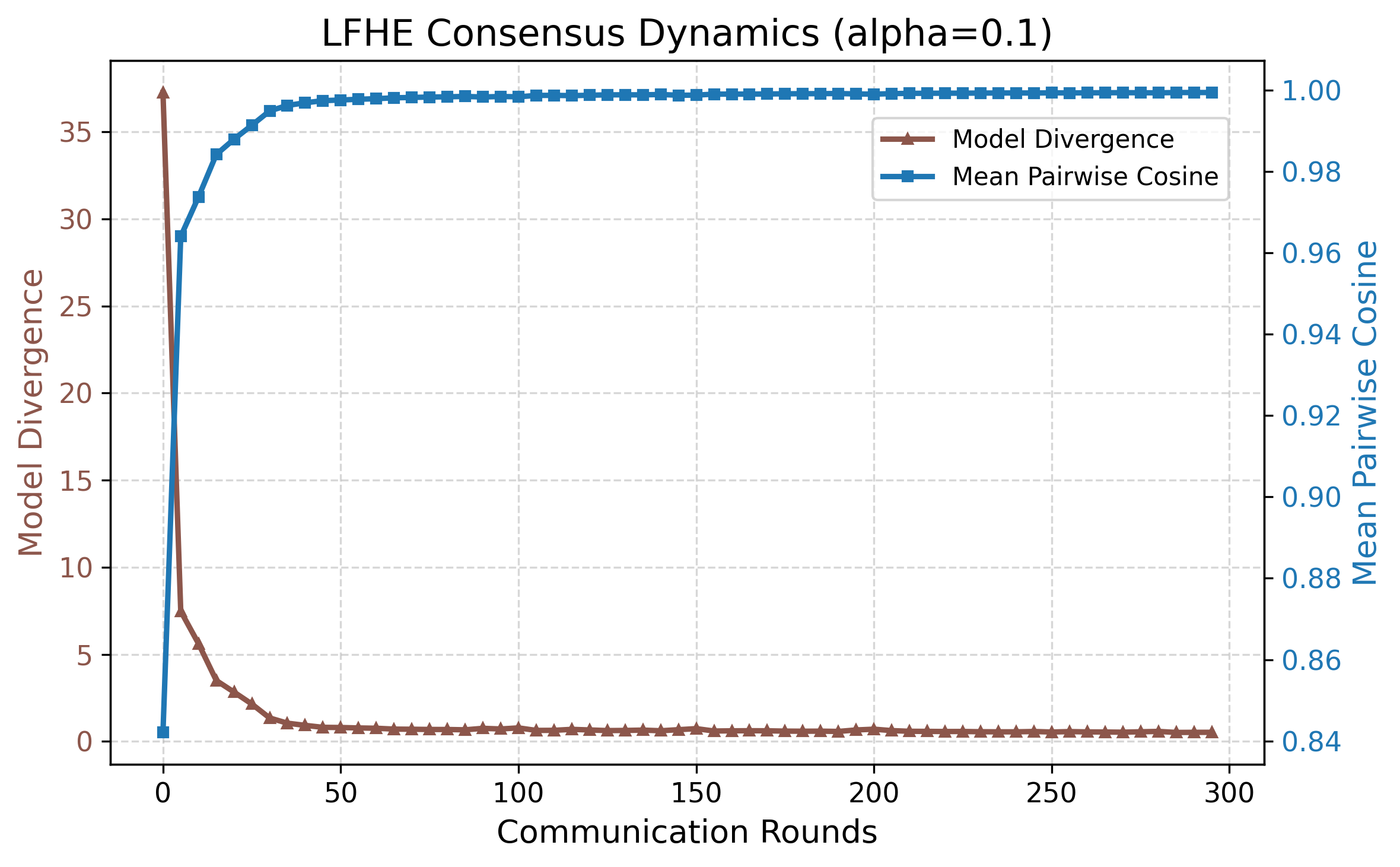}
    \caption{Consensus dynamics.}
    \label{fig:consensus_alpha}
\end{subfigure}
\hfill
\begin{subfigure}{0.48\linewidth}
    \centering
    \includegraphics[width=\linewidth]{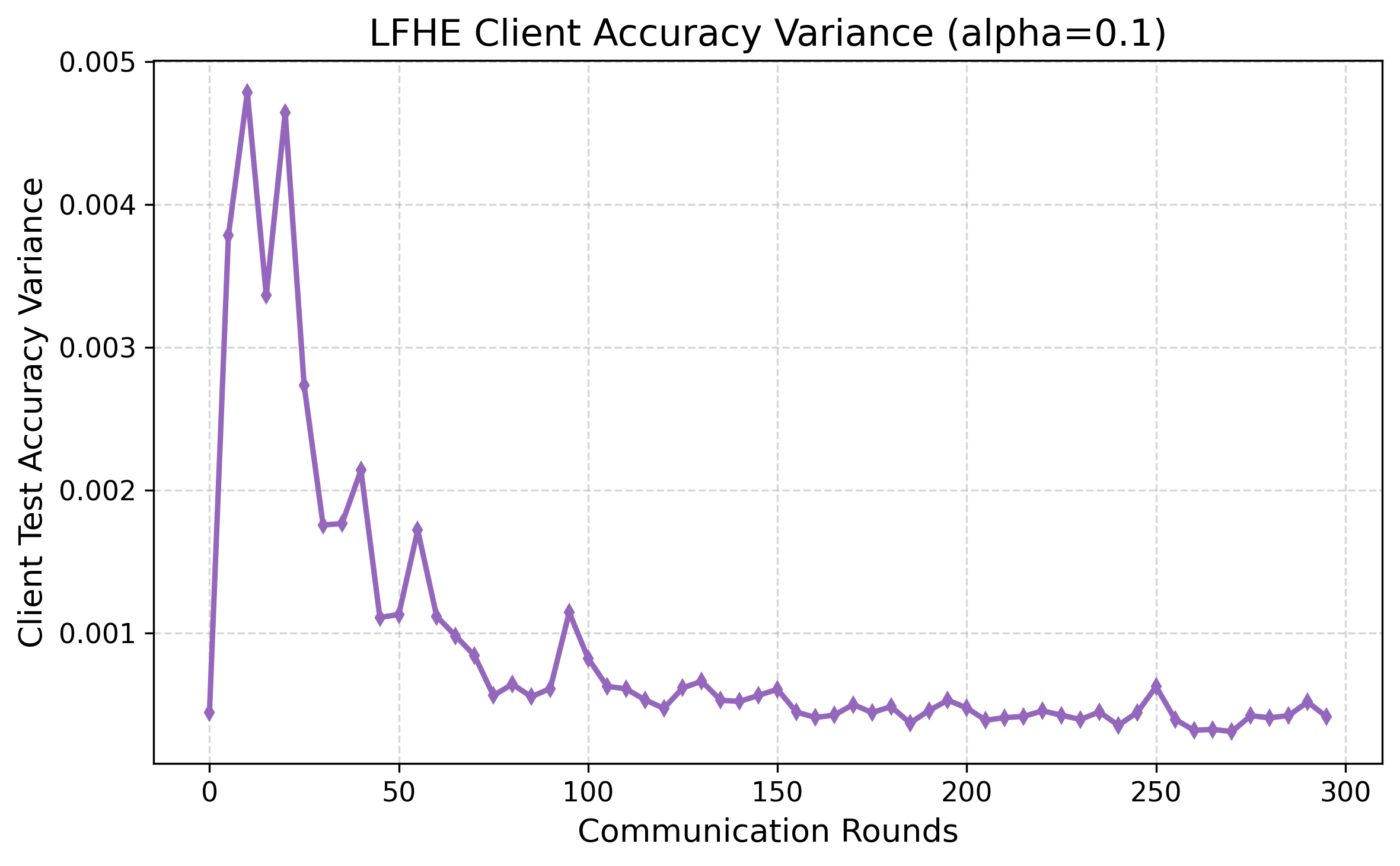}
    \caption{Client accuracy variance.}
    \label{fig:accvar_alpha}
\end{subfigure}
\caption{Additional LFHE diagnostics on CIFAR-10 with $\alpha=0.1$. Left: model-consensus quantities. Right: client-level test-accuracy variance.}
\label{fig:consensus_stability}
\end{figure}

Figure~\ref{fig:consensus_stability} reports model-consensus and client-accuracy-variance diagnostics for LFHE on CIFAR-10 with \(\alpha=0.1\). Model divergence decreases and mean pairwise cosine similarity rises during training; client-level accuracy variance also declines after the early phase. These are descriptive within-method trajectories. Without matched causal interventions, they should not be interpreted as proving that a particular topology statistic causes the learning improvement.

\section{Bridge Evidence Under Matched Rewiring Budget}
\label{app:bridge}

To more directly test whether the proposed structural term is associated with improved communication structure rather than topology churn alone, we conduct a controlled bridge experiment on CIFAR-10 under highly non-IID data (\(\alpha=0.1\)). The goal is not to re-evaluate the full learning system against all baselines, but to isolate how different local rewiring objectives affect topology quality and optimization behavior under a matched local update protocol.

\subsection{Protocol}

We compare five methods: \textbf{Static}, \textbf{Similarity-only}, \textbf{Exploration-only}, \textbf{Structural-only}, and \textbf{LFHE}. All adaptive variants use the same FoF candidate-generation rule, the same maximum rewiring budget per update, the same initial random graph, and the same training and aggregation protocol. Thus, the compared methods differ only in the local score used to rank candidate updates; candidate access, update frequency, and communication budget are held fixed.

The compared local rewiring scores are
\[
\text{Similarity-only: } S_i - \omega_3 \frac{\deg(i)}{D_{\max}},
\qquad
\text{Exploration-only: } D_i - \omega_3 \frac{\deg(i)}{D_{\max}},
\]
\[
\text{Structural-only: } C_i - \omega_3 \frac{\deg(i)}{D_{\max}},
\qquad
\text{LFHE: } \omega_1(1-\beta_t)C_i + \omega_2\beta_t D_i - \omega_3 \frac{\deg(i)}{D_{\max}}.
\]
All methods therefore operate under the same local rewiring mechanism and differ only in the objective used to evaluate candidate updates.

\subsection{Results}

\begin{figure*}[t]
\centering
\includegraphics[width=0.92\linewidth]{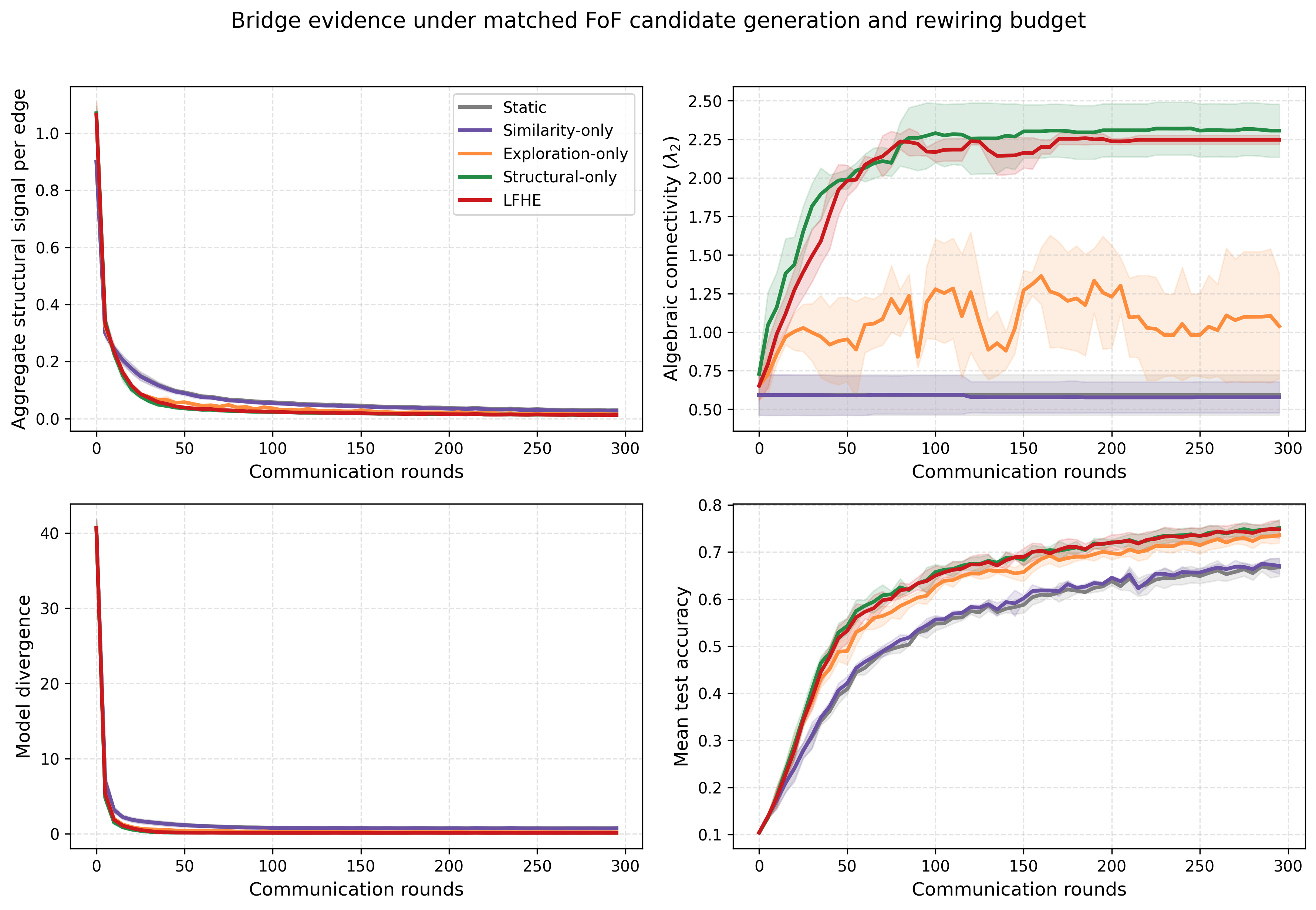}
\caption{\small Bridge evidence under matched FoF candidate generation and rewiring budget on CIFAR-10 with \(\alpha=0.1\). Methods containing the structural term (\textbf{Structural-only} and \textbf{LFHE}) produce substantially larger and more stable gains in algebraic connectivity than \textbf{Similarity-only} or \textbf{Exploration-only}. The same methods also achieve lower model divergence and stronger test accuracy under the same local update protocol.}
\label{fig:bridge_evidence}
\end{figure*}

\begin{figure}[t]
\centering
\includegraphics[width=0.82\linewidth]{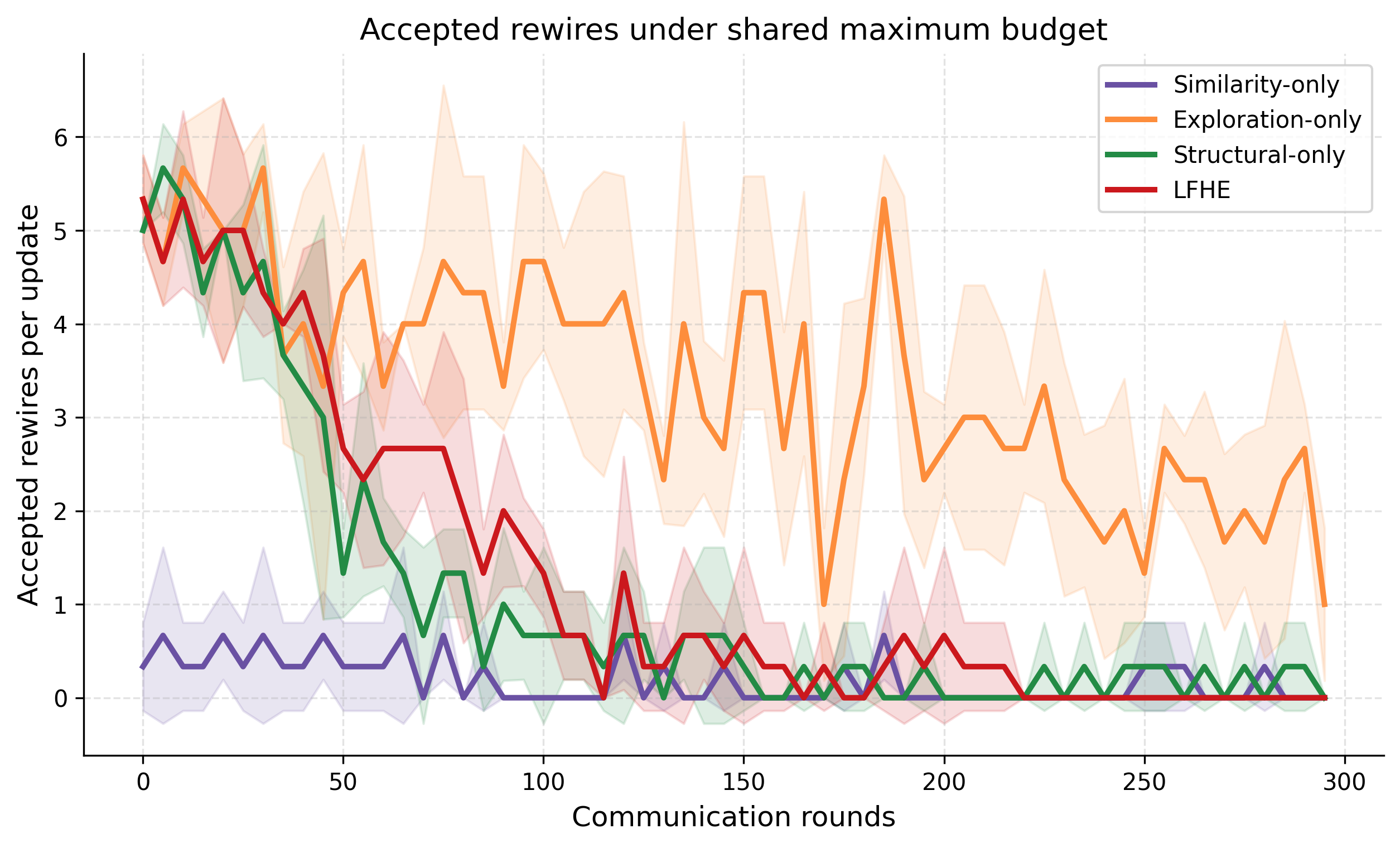}
\caption{\small Accepted rewires per topology update under the same maximum rewiring budget. \textbf{Exploration-only} performs more rewiring than \textbf{LFHE} or \textbf{Structural-only} during much of training, but does not achieve the strongest connectivity or accuracy. This shows that the advantage of structural-term methods is not explained by rewiring frequency alone.}
\label{fig:bridge_budget}
\end{figure}

Table~\ref{tab:bridge_summary} summarizes the matched-budget bridge evidence, while Figures~\ref{fig:bridge_evidence} and~\ref{fig:bridge_budget} show the full trajectories. Because all adaptive variants use the same candidate-generation rule, rewiring budget, and training protocol, the observed differences are attributable to the local rewiring objective rather than to protocol asymmetries.

\begin{table}[t]
\centering
\small
\caption{\small Summary of matched-budget bridge evidence on CIFAR-10 with \(\alpha=0.1\). Final \(\lambda_2\) reports the final algebraic connectivity. \(\Delta \lambda_2\) is computed relative to the shared initial graph used by all matched-budget variants. Mean divergence is computed over the last 20\% of evaluation rounds.}
\label{tab:bridge_summary}
\setlength{\tabcolsep}{4pt}
\begin{tabular}{lccccc}
\toprule
Method & Final \(\lambda_2\) & \(\Delta \lambda_2\) & Final Acc. & Mean Div. & Mean Rewires \\
\midrule
Static           & 0.592 & 0.000  & 0.6675 & 0.766 & 0.000 \\
Similarity-only  & 0.578 & -0.014 & 0.6704 & 0.727 & 0.183 \\
Exploration-only & 1.038 & 0.447  & 0.7356 & 0.221 & 3.406 \\
Structural-only  & 2.306 & 1.715  & 0.7508 & 0.136 & 1.111 \\
LFHE             & 2.247 & 1.655  & 0.7482 & 0.133 & 1.333 \\
\bottomrule
\end{tabular}
\end{table}

First, the two objectives containing the structural term (\textbf{Structural-only} and \textbf{LFHE}) produce the largest gains in the offline algebraic-connectivity diagnostic. Structural-only is slightly stronger than full LFHE in this regime: final \(\lambda_2\) is \(2.306\) versus \(2.247\), and final accuracy is \(0.7508\) versus \(0.7482\). This identifies the structural term as the main empirical signal in the matched protocol and rules out a claim that the complete objective is universally optimal.

Second, the connectivity gains are not explained by rewiring frequency alone. As shown in Table~\ref{tab:bridge_summary} and Figure~\ref{fig:bridge_budget}, \textbf{Exploration-only} performs the most accepted rewires on average (\(3.406\) per update), yet it does not achieve the strongest connectivity, the lowest model divergence, or the best final accuracy. Conversely, \textbf{Structural-only} and \textbf{LFHE} achieve substantially stronger topology and optimization outcomes with far fewer rewires. This indicates that the benefit comes from \emph{which} local objective guides rewiring, rather than from topology churn alone.

Taken together, these matched score controls show that topology churn alone does not explain the observed structure and that \(C_i\) is the dominant score component in this regime. The matched score-factor diagnostic isolates the effect of the local scoring rule under shared FoF access; the next subsection adds score-free Random-FoF, degree-preserving random-swap, and representation-free graph-only controls to test broader alternative explanations. None of these results implies that \(C_i(t)\) monotonically tracks \(\lambda_2\).

\subsection{Additional Simple Controls}

The score-factor study above is complemented by simpler controls designed to test alternative explanations. Across three seeds, a matched Random-FoF variant reaches \(74.62\pm2.25\%\) final accuracy, close to LFHE's \(74.82\%\) in this diagnostic, but has lower final algebraic connectivity (\(1.696\) versus \(2.247\)). On seed 42, Random-FoF commits 81 accepted operations and reaches 73.24\% final accuracy with 61.26\% nAUC. A degree-preserving random-swap diagnostic accepts 297 double-edge swaps but reaches only 66.24\% accuracy. Conversely, a representation-free graph-only control reaches a higher final \(\lambda_2\) of 2.403 but only 73.52\% accuracy.

These controls rule out two simple explanations. First, more topology activity is not sufficient for better learning. Second, maximizing a graph-only connectivity diagnostic is not sufficient either. The Random-FoF result is also a useful caution: its final accuracy is close to LFHE in the matched regime, so the evidence supports a structural difference more strongly than a large universal accuracy gap. We therefore avoid attributing all LFHE gains uniquely to its score.

\subsection{Interpretation}

This experiment complements the removal ablation by holding FoF access fixed and varying only the score. Together with Appendix~\ref{appendix:theory}, it separates two claims: \(C_i\) has an exact state-dependent Dirichlet-energy interpretation, and the matched experiment shows that using this signal produces the strongest measured topology change in this regime. Structural-only can still outperform the full objective, so exploration and degree control are best understood as search and resource regulators rather than theoretically necessary components of every optimal update.

\section{Cost Analysis}
\label{app:cost}

In this appendix, we analyze the computational overhead of LFHE-based topology optimization in the decentralized training pipeline. Our implementation records both wall-clock topology-update time and a set of control-overhead statistics, including neighbor-list reads, FoF candidate checks, fitness evaluations, accepted rewires, and edge changes. These quantities are accumulated during training and summarized across multiple random seeds.

\subsection{Measurement Protocol}

The cost analysis follows the main CIFAR-10 configuration with \(N=30\), \(T=300\), local epoch \(E=1\), topology interval \(K=5\), and online degree threshold \(D_{\max}=4\). LFHE is therefore executed every five communication rounds, resulting in \(60\) topology update calls per seed over the full training process. For each topology update, the implementation measures the elapsed wall-clock time of \textsc{LFHE-Update}:
\[
\tau_t = t_{\mathrm{end}} - t_{\mathrm{start}},
\]
where \(t_{\mathrm{start}}\) and \(t_{\mathrm{end}}\) are recorded immediately before and after the topology update routine. The total topology overhead and amortized per-round overhead are defined as
\[
\tau_{\mathrm{total}} = \sum_{t \in \mathcal{T}_{\mathrm{upd}}} \tau_t,
\qquad
\tau_{\mathrm{round}} = \frac{\tau_{\mathrm{total}}}{T},
\]
where \(\mathcal{T}_{\mathrm{upd}}\) denotes the set of update rounds.

In addition to runtime, we record several control-overhead statistics:
\begin{itemize}[nosep]
\item \textbf{Neighbor-list reads}: the number of accesses to local neighborhood structure;
\item \textbf{Candidate checks}: the number of FoF candidates examined;
\item \textbf{Fitness evaluations}: the number of node-level fitness computations;
\item \textbf{Accepted rewires}: the number of candidate rewiring operations that are committed;
\item \textbf{Edge changes}: the total number of edge insertions/removals between consecutive graph states. 
\end{itemize}

\subsection{Asymptotic Cost Discussion}

LFHE restricts discovery to ego-network and FoF information. Let \(\bar d=\max_i d_i\) denote the maximum degree actually present during an update, including any degree inherited from the initializer. The candidate-inspection cost is bounded by
\[
O\!\left(\sum_{i=1}^{N} d_i^2\right)
\subseteq O(N\bar d^2).
\]
If the initializer and online protocol keep \(\bar d\) bounded independently of \(N\), the control computation is linear in the number of clients. Because the reference initializer is specified by expected degree rather than a hard cap, this conditional statement is more precise than replacing \(\bar d\) unconditionally by \(D_{\max}\).

In addition, each candidate test only requires a local fitness computation based on the current representation of a node and its neighbors. Hence, LFHE avoids repeated global eigendecomposition or full-graph optimization during training.

\subsection{Observed Runtime Overhead}

Across seeds 42, 43, and 44, the mean total topology-update time is \(1.886 \pm 0.038\) seconds for the entire 300-round training run. When amortized over all communication rounds, this corresponds to only \(0.00629 \pm 0.00013\) seconds per round. Since topology evolution is triggered once every five communication rounds, the mean time per topology update is \(0.0314 \pm 0.0006\) seconds.

These measurements show small simulator-side topology-update time on the reported workstation. They do not include real network latency or all deployment control traffic, so they should be interpreted as implementation overhead for this synchronous simulator rather than an end-to-end systems cost.

\begin{figure}[t]
    \centering
    \includegraphics[width=0.75\linewidth]{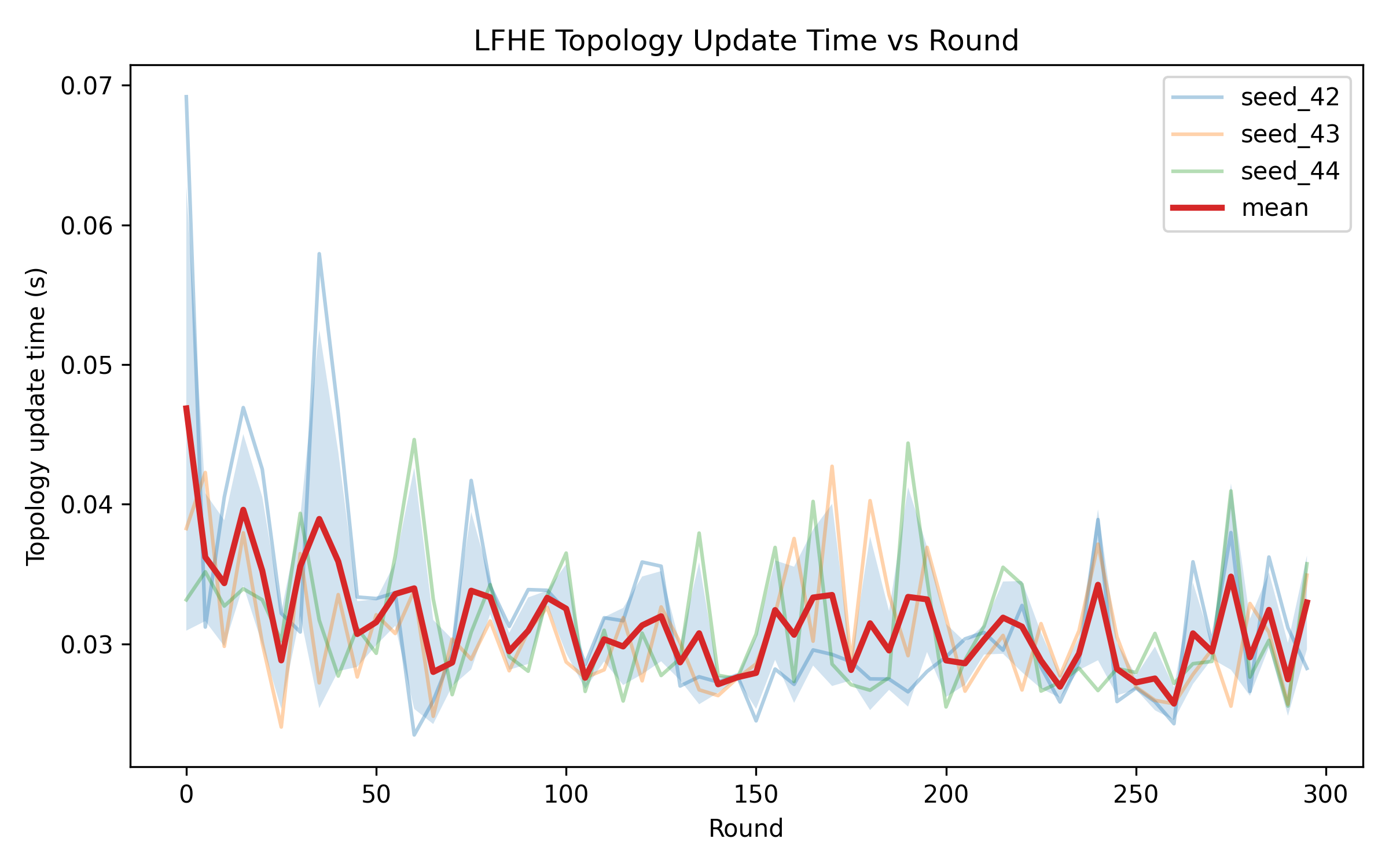}
    \caption{Per-update LFHE topology overhead across communication rounds. Thin lines show individual seeds and the bold line shows the mean across seeds. The runtime remains stable throughout training, indicating that LFHE does not incur increasing topology-update cost in later rounds.}
    \label{fig:cost_topology_time}
\end{figure}

As shown in Figure~\ref{fig:cost_topology_time}, the wall-clock cost of LFHE updates fluctuates slightly across rounds and seeds but remains within a narrow range. This suggests that the local rewiring mechanism does not become progressively more expensive as training proceeds.

\subsection{Control Overhead}

The control statistics reveal that LFHE performs bounded local exploration while committing only a small number of structural updates. Averaged across seeds, LFHE performs \(353.78 \pm 48.55\) candidate checks and the same number of fitness evaluations per update, while accepting only \(0.339 \pm 0.122\) rewires per update.

Over the full training run, this corresponds to \(21{,}226.67 \pm 2{,}913.07\) candidate checks, \(21{,}226.67 \pm 2{,}913.07\) fitness evaluations, and \(20.33 \pm 7.32\) accepted rewires. The total number of neighbor-list reads is \(8{,}326 \pm 562.75\), and the total number of edge changes is \(43.33 \pm 15.84\).

These results indicate that LFHE explores a large number of local candidates but commits only a very small subset of rewiring operations. This behavior is consistent with a conservative accept-if-improves policy, which favors stability over aggressive topology modification.

\begin{figure}[t]
    \centering
    \begin{subfigure}[t]{0.49\linewidth}
        \centering
        \includegraphics[width=\linewidth]{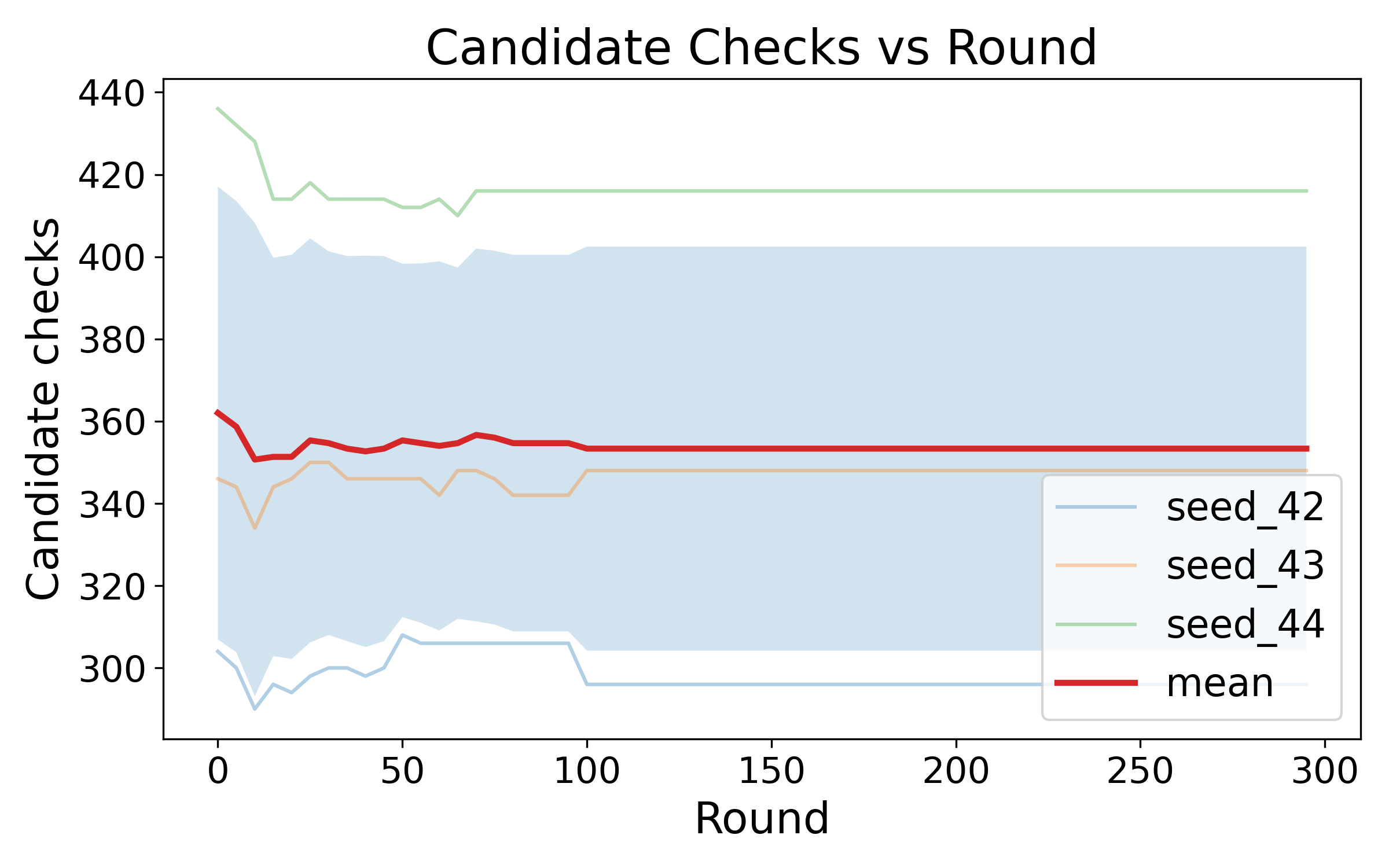}
        \caption{Candidate checks per topology update.}
        \label{fig:cost_candidate_checks}
    \end{subfigure}
    \hfill
    \begin{subfigure}[t]{0.49\linewidth}
        \centering
        \includegraphics[width=\linewidth]{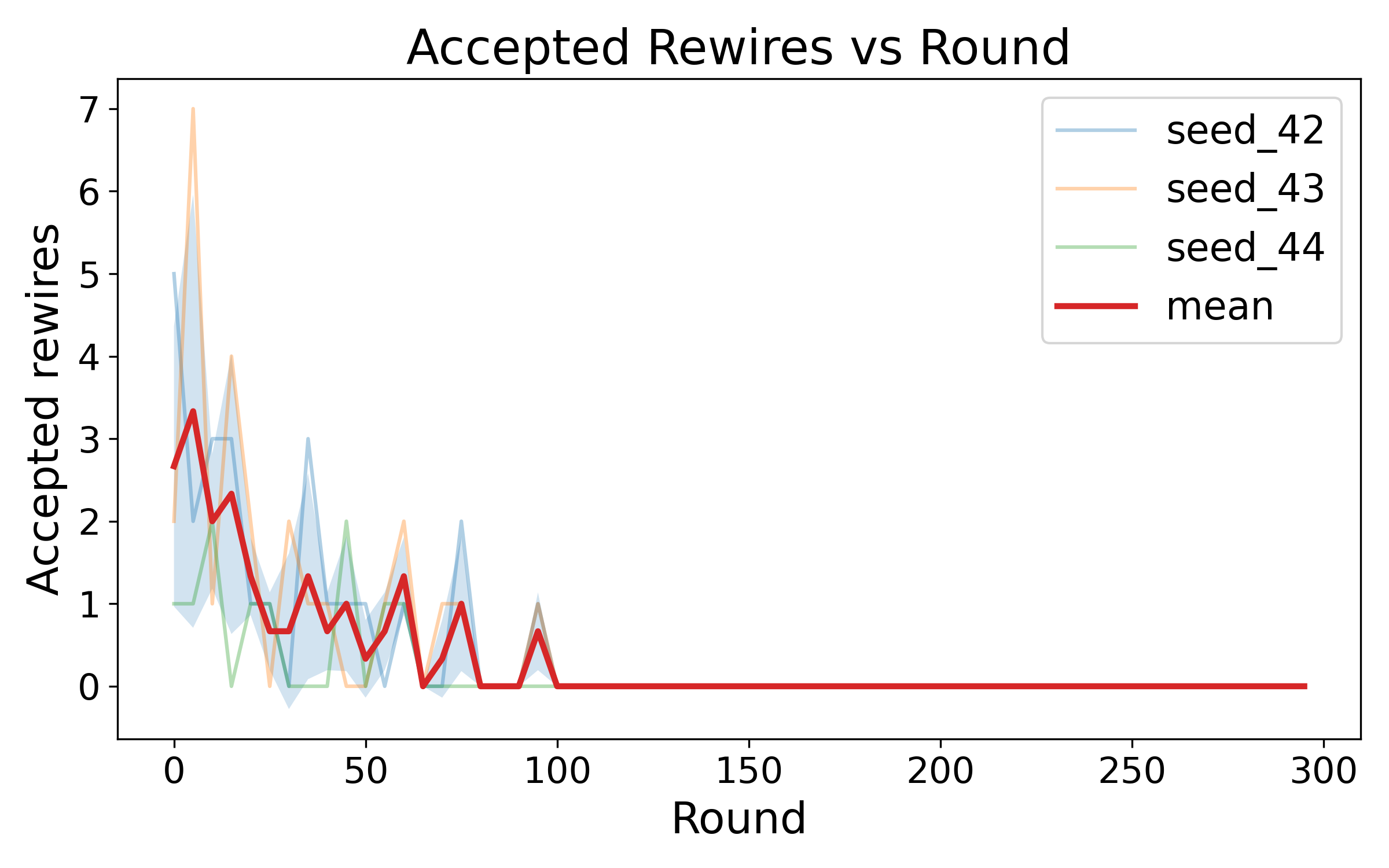}
        \caption{Accepted rewires per topology update.}
        \label{fig:cost_accepted_rewires}
    \end{subfigure}
    \caption{Control overhead of LFHE during topology evolution. \textbf{Left:} Candidate checks increase in the early stage and then stabilize, showing a stable candidate-check workload over later training rounds. \textbf{Right:} Accepted rewiring operations are concentrated in early rounds and become rare as the topology stabilizes.}
    \label{fig:cost_control_overhead}
\end{figure}

Figure~\ref{fig:cost_control_overhead} further illustrates this behavior. As shown in Figure~\ref{fig:cost_candidate_checks}, the number of candidate checks rises during the early phase of training and then stabilizes at around \(350\) checks per update. This shows that the recorded candidate-check workload remains approximately stable over later training rounds in these runs; it should not be interpreted as an empirical scaling law in \(N\). Figure~\ref{fig:cost_accepted_rewires} shows a complementary trend: accepted rewires are more frequent in the early stage but become rare in later rounds, consistent with a stabilizing communication topology.

\subsection{Acceptance Ratio}

A useful empirical measure of update selectivity is
\[
\mathrm{AcceptRate} =
\frac{\#\text{ accepted rewires}}{\max(\#\text{ candidate checks},1)}.
\]

\vspace{-5pt}

Using the full-run averages, this yields an efficiency on the order of \(10^{-3}\), indicating that only a very small fraction of evaluated candidates are accepted. Only a small fraction of inspected candidates is therefore committed. This is consistent with a conservative accept-if-improves rule at the initiator, but ``accepted'' should not be read as globally or bilaterally beneficial.

Taken together, Figures~\ref{fig:cost_candidate_checks} and~\ref{fig:cost_accepted_rewires} show that LFHE performs selective rather than aggressive topology modification: it continues to inspect local candidates, but commits only those rewiring operations that improve node-level fitness under the imposed constraints.

\end{document}